\documentclass[10pt,journal,compsoc]{IEEEtran}
\ifCLASSOPTIONcompsoc
  \usepackage[nocompress]{cite}
\else
  \usepackage{cite}
\fi
\ifCLASSINFOpdf
\else
\fi
\usepackage{lineno,hyperref}
\usepackage{algorithm}
\usepackage{algorithmicx}
\usepackage{amsmath,graphicx}
\usepackage{subfigure}
\usepackage{amsmath}
\usepackage{multirow}
\usepackage{cases}
\usepackage{subeqnarray}
\usepackage{xspace}
\usepackage{xcolor}
\usepackage{amsfonts}
\usepackage{amsthm}
\usepackage[noend]{algpseudocode}
\usepackage{diagbox}
\usepackage{booktabs}
\usepackage{enumitem}

\newcommand{\algorithmname}{Algorithm}
\newcommand{\equationname}{Eq.}
\newcommand{\method}{{FedAP}\xspace}

\begin{document}
%
\title{Personalized Federated Learning with Adaptive Batchnorm for Healthcare}
%
%
%
%

\author{Wang~Lu,
        Jindong~Wang,
        Yiqiang~Chen,~\IEEEmembership{Senior~Member,~IEEE,}
        Xin~Qin,
        Renjun~Xu,
        Dimitrios~Dimitriadis,~\IEEEmembership{Senior~Member,~IEEE,}
        and~Tao~Qin,~\IEEEmembership{Senior~Member,~IEEE}
\IEEEcompsocitemizethanks{
\IEEEcompsocthanksitem Wang Lu and Xin Qin are with University of Chinese Academy of Sciences and Institute of Computing Technology, Chinese Academy of Sciences.
E-mail: \{luwang, qinxin18b\}@ict.ac.cn

\IEEEcompsocthanksitem Jindong Wang and Tao Qin are with Microsoft Research Asia.\protect\\
E-mail: \{jindong.wang, taoqin\}@microsoft.com

\IEEEcompsocthanksitem Yiqiang Chen is with Institute of Computing Technology (CAS) and Pengcheng Laboratory. 
\protect
E-mail: yqchen@ict.ac.cn

\IEEEcompsocthanksitem Renjun Xu is with Zhejiang University. E-mail: rux@zju.edu.cn

\IEEEcompsocthanksitem Dimitrios Dimitriadis is with Microsoft Research. \protect\\
E-mail: didimit@microsoft.com

\IEEEcompsocthanksitem Correspondence to Jindong Wang and Yiqiang Chen.
}
\thanks{Manuscript received May 19, 2022.}
}

\markboth{Journal of \LaTeX\ Class Files,~Vol.~14, No.~8, April~2022}%
{Shell \MakeLowercase{\textit{et al.}}: Bare Demo of IEEEtran.cls for Computer Society Journals}
%



\IEEEtitleabstractindextext{%
\begin{abstract}
There is a growing interest in applying machine learning techniques to healthcare. Recently, federated learning (FL) is gaining popularity since it allows researchers to train powerful models without compromising data privacy and security. However, the performance of existing FL approaches often deteriorates when encountering non-iid situations where there exist distribution gaps among clients, and few previous efforts focus on personalization in healthcare. In this article, we propose \method to tackle domain shifts and then obtain personalized models for local clients. \method learns the similarity between clients based on the statistics of the batch normalization layers while preserving the specificity of each client with different local batch normalization. Comprehensive experiments on five healthcare benchmarks demonstrate that \method achieves better accuracy compared to state-of-the-art methods (e.g., \textbf{10}\%+ accuracy improvement for PAMAP2) with faster convergence speed.
\end{abstract}

}

\maketitle

\IEEEdisplaynontitleabstractindextext

%
\IEEEpeerreviewmaketitle

\IEEEraisesectionheading{\section{Introduction}\label{sec:introduction}}

%
%
%
%
\IEEEPARstart{M}{achine} learning has been widely adopted in many applications in people's daily life~\cite{sharma2021recurrent,brattoli2020rethinking,lu2022local}.
Specifically for healthcare, researchers can build models to predict health status by leveraging health-related data, such as activity sensors~\cite{lu2021cross}, images~\cite{fei2020deep}, and other health information~\cite{choi2018mime, unal2021escaped, lu2022semantic}.
To achieve satisfying performance, machine learning healthcare applications often require sufficient client data for model training.
However, with the increasing awareness of privacy and security, more governments and organizations enforce the protection of personal data via different regulations~\cite{inkster2018china,voigt2017eu}.
In this situation, federated learning (FL)~\cite{yang2019federated} emerges to build powerful machine learning models with data privacy well-protected.

\begin{figure}[t!]
	\centering
	\includegraphics[width=0.5\textwidth]{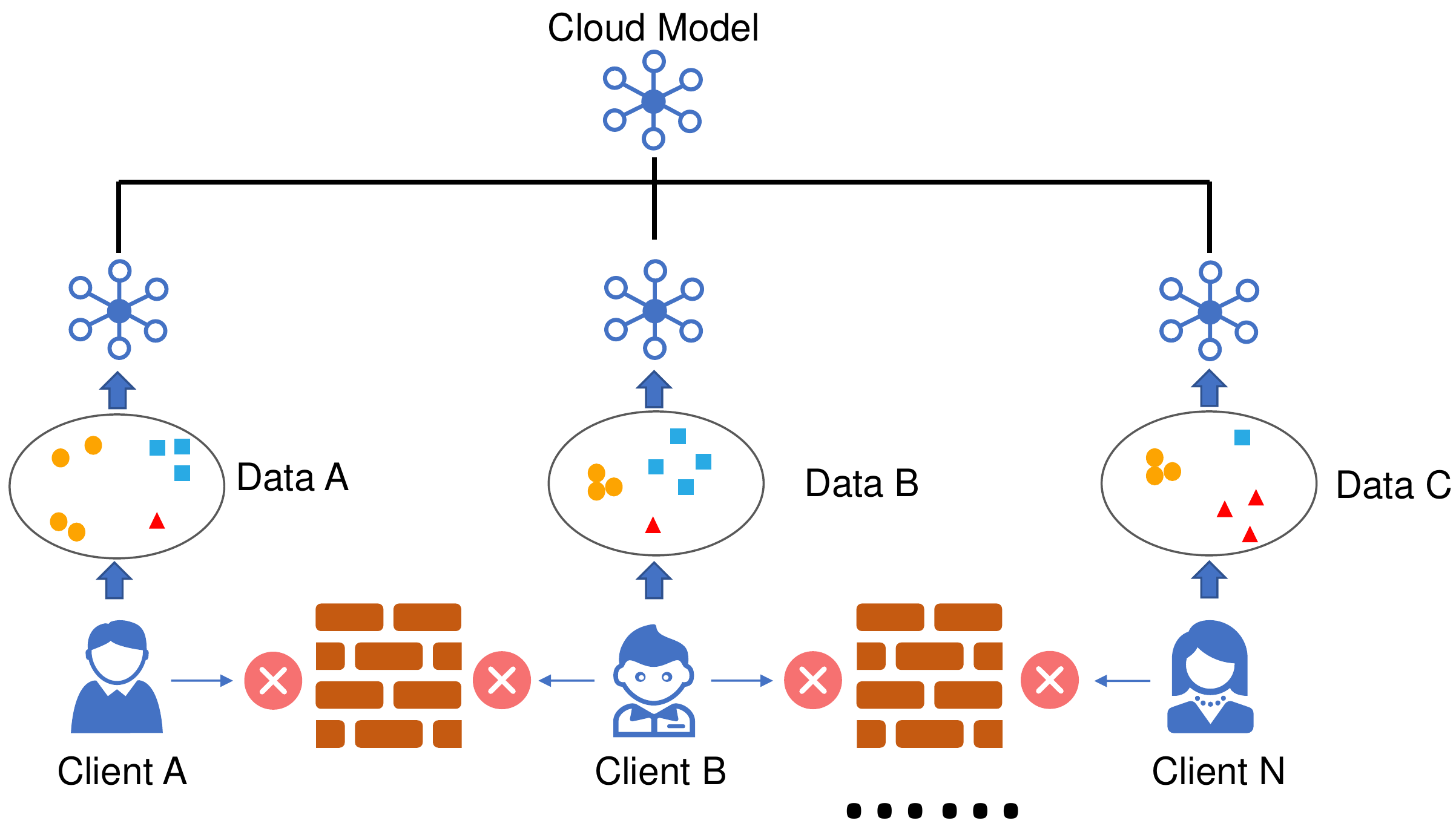}
	\caption{Data non-iid in federated learning: different clients have different data distributions.}
	\label{fig:fedavg}
\end{figure}

Personalization is important in healthcare applications since different individuals, hospitals or countries usually have different demographics, lifestyles, and other health-related characteristics~\cite{xu2021federated}, i.e., the non-iid issue (not identically and independently distributed).
Therefore, we are more interested in achieving better personalized healthcare, i.e., building FL models for each client to preserve their specific information while harnessing their commonalities.
As shown in \figurename~\ref{fig:fedavg}, there are three different clients $A, B$, and $C$ with different statistics of data distributions (e.g., the adult $A$ and the child $B$ may have different lifestyles and activity patterns).
Even if federated learning can perform in the standard way, the non-iid issue cannot be easily handled.
This will severely limit the performance of existing federated learning algorithms.

The popular FL algorithm, FedAvg~\cite{mcmahan2017communication}, has demonstrated superior performance in many situations~\cite{ching2018opportunities,zhu2020application}.
However, FedAvg is unable to deal with non-iid data among different clients since it directly averages the parameters of models coming from all participating clients~\cite{li2019convergence}.
There are some algorithms for this non-iid situation.
FedProx~\cite{li2018federated} is designed for non-iid data. However, FedProx only learns a global model for all clients, which means that it is unable to obtain personalized models for clients.
FedHealth~\cite{chen2020fedhealth}, another work for personalized healthcare, needs access to a large public dataset, which is often impossible in real applications.
FedBN~\cite{li2021fedbn} handles the non-iid issue by learning local batch normalization layers for each client but ignores the similarities across clients that can be used to boost the personalization.

In this article, we propose \method, a personalized federated learning algorithm via \emph{adaptive batch normalization} for healthcare.
Specifically, \method learns the similarities among clients with the help of a pre-trained model that is easy to obtain.
The similarities are determined by the distances of the data distributions, which can be calculated via the statistical values of the layers' outputs of the pre-trained network.
After obtaining the similarities, the server averages the models' parameters in a personalized manner and generates a unique model for each client.
Each client preserves its own batch normalization and updates the model with a momentum method.
In this way, \method can cope with the non-iid issue in federated learning.
\method is extensible and can be deployed to many healthcare applications.

Our contributions are as follows:
\begin{enumerate}
    \item We propose \method, a personalized federated learning algorithm via adaptive batch normalization for healthcare, which can aggregate the information from different clients without compromising privacy and security, and learn personalized models for each client.
    \item We evaluate the performance of \method in five public healthcare datasets across time series and image modalities. Experiments demonstrate that our \method achieves significantly better performance than state-of-the-art methods in all datasets.
    \item \method reduces the number of rounds and speeds up the convergence to some extent. Moreover, some experimental results illustrate \method may be able to reduce communication costs with little performance degradation via increasing local iterations and decreasing global communications.
\end{enumerate}

\section{Related Work}
\subsection{Machine Learning and Healthcare}
With the rapid development of perception and computing technology, people can make use of machine learning to help doctors diagnose~\cite{chen2021diagnose} and assist doctors in the operation~\cite{avellino2019impacts}, etc. 
Many methods are proposed to monitor people's health state~\cite{muhammad2020eeg} and diagnose diseases that may even have better performance than doctors', especially in the field of medical images~\cite{ji2021learning}.
Moreover, machine learning can make disease warnings via daily behavior supervision with simple wearable sensors~\cite{sun2020gait}. 
For instance, certain activities in daily life reflect early signals of some cognitive diseases. 
Through daily observation of gait changes and finger flexibility, the machine can tell people whether they are suffering from Parkinson~\cite{chen2017pdassist}. 
In addition, some studies worked for better personalization in healthcare~\cite{vogenberg2010personalized, chung2020deep}.

Unfortunately, a successful healthcare application needs a large amount of labeled data of persons. 
However, in real applications, data are often separate and few people or organizations are willing to disclose their private data. 
In addition, an increasing number of regulations, such as~\cite{inkster2018china,voigt2017eu}, hold back the leakages of data.
These make different clients cannot exchange data directly, and the scattered data forms separate data islands, which makes it impossible to learn a traditional model with aggregated data.

\subsection{Federated Learning}
Federated learning is a usual way to combine each client's information while protecting data privacy and security~\cite{yang2019federated}. 
It was first proposed by Google~\cite{mcmahan2017communication}, where they proposed FedAvg to train machine learning models via aggregating distributed mobile phones' information without exchanging data. 
The key idea is to replace direct data exchanges with model parameter-related exchanges. 
FedAvg is able to resolve the data islanding problems.

Although federated learning is an emerging field, it has attracted much attention~\cite{abdulrahman2020survey, zhang2021survey}. 
Federated learning can be divided into horizontal federated learning, vertical federated learning, and federated transfer learning according to the characteristics data. 
When the client features of the two datasets overlap a lot but the clients overlap little, horizontal federated learning can be applied~\cite{mcmahan2017communication}.
In the horizontal federated learning, datasets are split horizontally and the clients share the same features finally.
For example, Smith et al.~\cite{smith2017federated} proposed a novel systems-aware optimization method, MOCHA, to solve security problems in multitasking.
When the client features of the two datasets overlap little but the clients overlap a lot, we can utilize vertical federated learning, where different clients have different columns of the features~\cite{hardy2017private}. 
For example, Cheng et al.~\cite{cheng2021secureboost} proposed a novel lossless privacy-preserving tree-boosting system known as SecureBoost to jointly conduct over multiple parties with partially common client samples but different feature sets.
When the clients and client features of the two datasets both rarely overlap, federated transfer learning is often utilized~\cite{liu2020secure, he2020group}.
For example, Yoon et al.~\cite{yoon2021federated} proposed a novel federated continual learning framework, Federated Weighted Inter-client Transfer (FedWeIT), which decomposed the network weights into global federated parameters and sparse task-specific parameters.
In~\cite{yoon2021federated}, each client received selective knowledge from other clients by taking a weighted combination of their task-specific parameters.
In addition, many methods, such as differential privacy, are proposed to protect data further~\cite{dwork2008differential, agarwal2018cpsgd}.
In this paper, we mainly focus on horizontal federated learning when the training data are not independent and identically distributed (Non-IID) on the clients.

\begin{figure*}[t!]
	\centering
	\includegraphics[width=0.9\textwidth]{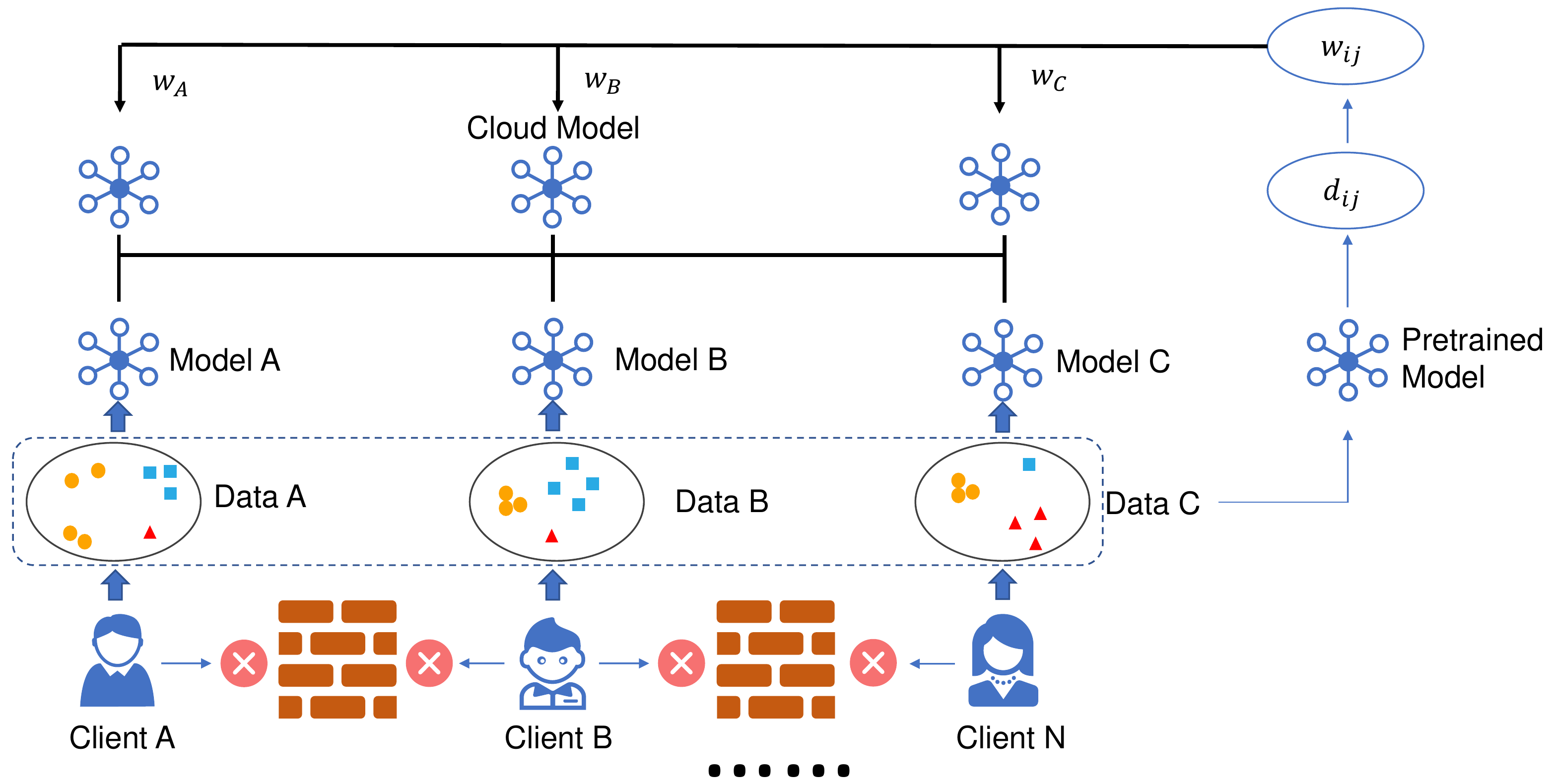}
	\caption{The structure of the \method method.}
	\label{fig:wfedbn}
\end{figure*}

Although FedAvg works well in many situations, it may still suffer from the non-iid data and fail to build personalized models for each client~\cite{smith2017federated, khodak2019adaptive, zhu2021data}. 
A survey about federated Learning on non-iid Data can be found here~\cite{zhu2021federated}.
FedProx~\cite{li2018federated} tackled data non-iid by allowing partial information aggregation and adding a proximal term to FedAvg. 
\cite{yeganeh2020inverse} aggregated the models of the clients with weights computed via $L_1$ distance among client models' parameters. 
These works focus on a common model shared by all clients while some other works try to obtain a unique model for each client. 
\cite{arivazhagan2019federated} exchanged information of base layers and preserved personalization layer to combat the ill-effects of non-iid. 
\cite{dinh2020personalized} utilized Moreau envelopes as clients’ regularized loss function and decoupled personalized model optimization from the global model learning in a bi-level problem stylized for personalized FL. 
\cite{yu2020salvaging} evaluated three techniques for local adaptation of federated models: fine-tuning, multi-task learning, and knowledge distillation.
\cite{mansour2020three} also proposed and analyzed three approaches: user clustering, data interpolation, and model interpolation.
\cite{liang2020think} tried to jointly learn compact local representations on each device and a global model across all devices with a theoretic analysis.
\cite{deng2020adaptive} proposed APFL where each client would train their
local models while contributing to the global model. 
Another work~\cite{sattler2020clustered}, Clustered Federated Learning (CFL), grouped
the client population into clusters with jointly trainable data distributions.
Two works most relevant to our method are FedHealth~\cite{chen2020fedhealth} and FedBN~\cite{li2021fedbn}. 
FedHealth needs to share some datasets with all clients while FedBN used local batch normalization to alleviate the feature shift before averaging models. 
Although there are already some works to cope with data non-iid, few works pay attention to feature shift non-iid and other shifts at the same time and obtaining an individual model for each client in healthcare.

\subsection{Batch Normalization}
Batch Normalization (BN)~\cite{ioffe2015batch} is an important component of deep learning. 
Batch Normalization improves the performance of the model and has a natural advantage in dealing with domain shifts. 
Li et al.~\cite{li2018adaptive} proposed an adaptive BN for domain adaptation where they learned domain-specific BN layers. 
Nowadays, researchers have explored many effects of BN, especially in transfer learning~\cite{segu2020batch}. 
FedBN~\cite{li2021fedbn} is one of few applications of BN in the field of FL field. 
However, FedBN does still not make full use of BN properties, and it does not consider the similarities among the clients.

\section{Method}

\subsection{Problem Formulation}

In federated learning, there are $N$ different clients (organizations or users), denoted as $\{ C_1, C_2, \cdots, C_N \}$ and each client has its own dataset, i.e. $\{ \mathcal{D}_1, \mathcal{D}_2, \cdots, \mathcal{D}_N \}$. 
Each dataset $\mathcal{D}_i = \{ (\mathbf{x}_{i,j}, y_{i,j}) \}_{j=1}^{n_i}$ contains two parts, i.e. a train dataset $\mathcal{D}_i^{tr} = \{ (\mathbf{x}_{i,j}^{tr}, y_{i,j}^{tr}) \}_{j=1}^{n_i^{tr}}$ and a test dataset $\mathcal{D}_i^{te} = \{ (\mathbf{x}_{i,j}^{te}, y_{i,j}^{te}) \}_{j=1}^{n_i^{te}}$.
Obviously, $n_i = n_i^{tr} + n_i^{te}$ and $\mathcal{D}_i = \mathcal{D}_i^{tr} \cup \mathcal{D}_i^{te}$. 
All of the datasets have different distributions, i.e. $P(\mathcal{D}_i) \neq P(\mathcal{D}_j)$.
Each client has its own model denoted as $\{ f_i\}_{i=1}^N$. 
Our goal is to aggregate information of all clients to learn a good model $f_i$ for each client on its local dataset $\mathcal{D}_i$ without private data leakage:
\begin{equation}
    \min_{\{f_k\}_{k=1}^N} \frac{1}{N} \sum_{i=1}^N \frac{1}{n_{i}^{te}} \sum_{j=1}^{n_i^{te}} \ell(f_i(\mathbf{x}_{i,j}^{te}), y_{i,j}^{te}),
    \label{eqa:goal}
\end{equation}
where $\ell$ is a loss function.

\subsection{Motivation}
There are mainly two challenges for personalized healthcare: data islanding and personalization. 
Following FedAvg~\cite{mcmahan2017communication} and some other traditional federated learning methods~\cite{gao2021convergence,cao2021provably}, it is easy to cope with the first challenge.
Personalization is a must in many applications, especially in healthcare.
It is better to train a unique model in each client for personalization.
However, one client often lacks enough data to train a model with high accuracy in federated learning.
In addition, clients do not have access to the data of other clients.
Overall, it is a challenge that how to achieve personalization to obtain high accuracy in federated learning.
As mentioned in~\cite{li2018adaptive}, batch normalization (BN) layers contain sufficient statistics (including mean and standard deviation) of features (outputs of layers). 
Therefore, BN has been utilized to represent distributions of training data indirectly in many works~\cite{li2018adaptive,chang2019domain}. 
We mainly use BN to represent the distributions of clients. 
Therefore, on the one hand, we utilize local BN to preserve clients' feature distributions. 
On the other hand, we also use BN-related statistics to calculate the similarity between clients for better personalization with weighted aggregation\footnote{Please note that we only share the statistics of the batch normalization layers, and no existing work shows that any methods can restore the specific sample with the statistics of certain layers~\cite{zhu2019deep,yin2021see}.}.

\begin{figure*}[t!]
	\centering
	\includegraphics[width=0.9\textwidth]{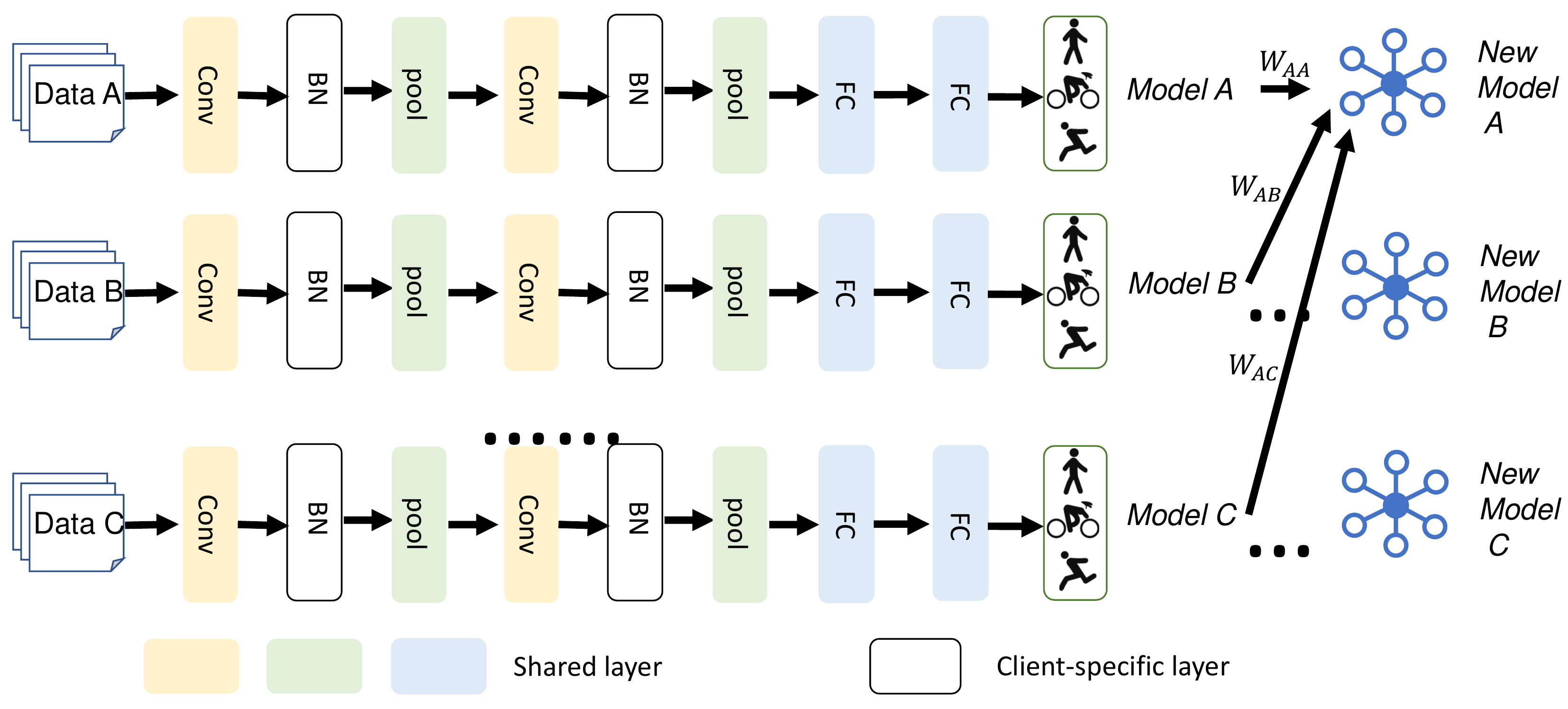}
	\caption{The concrete process of the \method.}
	\label{fig:fedbn}
\end{figure*}

\subsection{Our Approach: \method}

In this paper, we propose \method (Adaptive Federated Learning) to achieve accurate personal healthcare via adaptive batch normalization without compromising data privacy and security.
\figurename~\ref{fig:wfedbn} gives an overview of its structure.
Without loss of generality, we assume there are three clients, which can be extended to more general cases. 
The structure mainly contains five steps:
\begin{enumerate}
    \item The server distributes the pre-trained model to each client.
    \item Each client computes statistics of the outputs of specific layers according to local data.
    \item The server obtains the client similarities denoted by weight matrix $\mathbf{W}$ to guide aggregation.
    \item Each client updates its own model with the local train data and pushes its model to the server.
    \item The server aggregates models and obtains $N$ models delivered to $N$ clients respectively.
\end{enumerate}

For stability and simplicity, we only calculate $\mathbf{W}$ once and we show that computing once is enough to achieve acceptable performance in experiments.
Note that all processes do not involve the direct transmission of data, so \method avoids the leakage of private data and ensures security.
The keys of \method are obtaining $\mathbf{W}$ and aggregating the models.
We will introduce how to compute $\mathbf{W}$ after describing the process of model aggregation.

We denote the parameters of each model $f_i$ as $\boldsymbol{\theta}_i = \boldsymbol{\phi}_i \cup \boldsymbol{\psi}_i$, where $\boldsymbol{\phi}_i$ corresponds to the parameters of BN layers specific to each client and $\boldsymbol{\psi}_i$ is the parameters of the other layers (colored blocks in \figurename~\ref{fig:fedbn}).
$\mathbf{W}$ is an $N \times N$ matrix, which describes the similarities among the clients. $w_{ij} \in [0,1]$ represents the similarity between client $i$ and client $j$: the larger $w_{ij}$ is, the more similar the two clients are. 

\figurename~\ref{fig:fedbn} demonstrates the process of model aggregation.
As shown in \figurename~\ref{fig:fedbn}, $\boldsymbol{\phi}_i$ is particular while $\boldsymbol{\psi}_i$ is computed according to $\mathbf{w}_{i}$, where $\mathbf{w}_{i}$ means the $i-$th row of $\mathbf{W}$, and $\boldsymbol{\psi}$, where $\boldsymbol{\psi} = \{ \boldsymbol{\psi}_i \}_{i=1}^N$. 
$\boldsymbol{\phi}_i$ is BN parameters that are not shared across clients while $\boldsymbol{\psi}_i$ is other parameters that are shared.
Let $\boldsymbol{\theta}_i^t = \boldsymbol{\phi}_i^t \cup \boldsymbol{\psi}_i^t $ represent the parameters of the model from client $i$ in the round $t$. 
After updating $\boldsymbol{\theta}_i^t$ with the local data from the $i-$th client, we obtain updated parameters $\boldsymbol{\theta}_i^{t*} = \boldsymbol{\phi}_i^{t*} \cup \boldsymbol{\psi}_i^{t*}$.
We use the $*$ notation to denote updated parameters.
Then, for aggregation on the server, we have the following updating strategy:
\begin{equation}
    \begin{cases}
    \boldsymbol{\phi}_i^{t+1}  = & \boldsymbol{\phi}_i^{t*}\\
    \boldsymbol{\psi}_i^{t+1}  = & \sum_{j=1}^N w_{ij}\boldsymbol{\psi}_j^{t*}.
    \end{cases}
    \label{eqa:updaterule}
\end{equation}

The overall process of \method is described in \algorithmname~\ref{alg:algorithm}.
In the next sections, we will introduce how to compute the weight matrix $\mathbf{W}$.

\begin{algorithm}[tb]
\caption{\method}
\label{alg:algorithm}
\textbf{Input}: A pre-trained model $f$, data of $N$ clients $\{\mathcal{D}_i\}_{i=1}^N$, $\lambda$\\
\textbf{Output}: Client models $\{f_i\}_{i=1}^N$
\begin{algorithmic}[1] 
\State Distribute $f$ to each client
\State Each client computes its statistics $(\boldsymbol{\mu}_i, \boldsymbol{\sigma}_i)$, where $\boldsymbol{\mu}_i$ represents the mean values while $\boldsymbol{\sigma}_i$ represents the covariance matrices. Push $(\boldsymbol{\mu}_i, \boldsymbol{\sigma}_i)$ to the server
\State Compute $\mathbf{W}$ according to the statistics
\State Update clients' model with local data. Push updated parameter $\{ \boldsymbol{\theta}_i^{t*} \}_{i=1}^N$ to the server
\State Update $\{ \boldsymbol{\theta}_i^{t+1} \}_{i=1}^N$ according to \equationname~\eqref{eqa:updaterule} and distribute them to the corresponding clients \State Repeat steps $4 \sim 5$ until convergence or maximum round reached
\end{algorithmic}
\end{algorithm}

\subsection{Evaluate Weights}

In this section, we will evaluate the weights with a pre-trained model $f$ and propose two alternatives to compute the weights. We mainly rely on the feature output statistics of clients' data in the pre-trained network to compute weights.

We denote with $l \in \{1, 2, \cdots, L \}$ in superscript notations the different batch normalization layers in the model. And $\mathbf{z}^{i,l}$ represents the input of $l-$th batch normalization layer in the $i-$th client. The input of the classification layer in the $i-$th client is denoted as $\mathbf{z}^i$ which represents the domain features. We assume $\mathbf{z}^{i,l}$ is a matrix, $\mathbf{z}^{i,l}_{c_{i,l}\times s_{i,l}}$ where $c_{i,l}$ corresponds to the channel number while $s_{i,l}$ is the product of the other dimensions. Similarly, $\mathbf{z}^i = \mathbf{z}^i_{c_i \times s_i}$. We feed $\mathcal{D}_i$ into $f$, and we can obtain $\mathbf{z}^{i,l}_{c_{i,l}\times s_{i,l}}$. Obviously, $s_{i,l} = e \times n_i$ where $e$ is an integer. Now, we try to compute statistics on the channels, and we treat $\mathbf{z}^{i,l}$ as a Gaussian distribution. For the $l-$th layer of the $i-$th client, it is easy to obtain its distribution, $\mathcal{N}(\boldsymbol{\mu}^{i,l}, \boldsymbol{\sigma}^{i,l})$. We only compute statistics of inputs of BN layers. And the BN statistics of the $i-$th client is formulated as:
\begin{equation}
    (\boldsymbol{\mu}_i,\boldsymbol{\sigma}_i) = [ (\boldsymbol{\mu}^{i,1},\boldsymbol{\sigma}^{i,1}), (\boldsymbol{\mu}^{i,2},\boldsymbol{\sigma}^{i,2}), \cdots, 
(\boldsymbol{\mu}^{i,L},\boldsymbol{\sigma}^{i,L})].
\label{eqa:zl}
\end{equation}

Now we can calculate the similarity between two clients.
It is popular to adopt the Wasserstein distance to calculate the distance between two Gaussian distributions: \begin{equation}
\begin{aligned}
    &W_2^2(\mathcal{N}(\boldsymbol{\mu}^{i,l},\boldsymbol{\sigma}^{i,l}), \mathcal{N}(\boldsymbol{\mu}^{j,l},\boldsymbol{\sigma}^{j,l}))\\ 
    = &||\boldsymbol{\mu}^{i,l} - \boldsymbol{\mu}^{j,l}||^2 +\\
    & tr(\boldsymbol{\sigma}^{i,l} + \boldsymbol{\sigma}^{j,l} - 2((\boldsymbol{\sigma}^{i,l})^{1/2}\boldsymbol{\sigma}^{j,l}(\boldsymbol{\sigma}^{i,l})^{1/2} )^{1/2}),
\end{aligned}
    \label{eqa:W}
\end{equation}  
where $tr$ is the trace of the matrix.
Obviously, it is costly and difficult to perform efficient calculations.
Similar to BN, we perform approximations and consider that each channel is independent of the others.
Therefore, $\boldsymbol{\sigma}^{i,l}$ is a diagonal matrix, i.e. $\boldsymbol{\sigma}^{i,l} = Diag(\mathbf{r}^{i,l})$.
Therefore, we compute the approximation of Wasserstein distance as:
\begin{equation}
\begin{aligned}
    &W_2^2(\mathcal{N}(\boldsymbol{\mu}^{i,l},\boldsymbol{\sigma}^{i,l}), \mathcal{N}(\boldsymbol{\mu}^{j,l},\boldsymbol{\sigma}^{j,l}))\\ 
    = &||\boldsymbol{\mu}^{i,l} - \boldsymbol{\mu}^{j,l}||^2 +
    ||\sqrt{\mathbf{r}^{i,l}} - \sqrt{\mathbf{r}^{j,l}}||_2^2.
\end{aligned}
    \label{eqa:Wapprox}
\end{equation}  

Thus, the distance between two clients $i, j$ is computed as:
\begin{equation}
\begin{aligned}
    d_{i,j}&= \sum_{l=1}^L W_2(\mathcal{N}(\boldsymbol{\mu}^{i,l},\boldsymbol{\sigma}^{i,l}), \mathcal{N}(\boldsymbol{\mu}^{j,l},\boldsymbol{\sigma}^{j,l}))\\ 
    &= \sum_{l=1}^L (||\boldsymbol{\mu}^{i,l} - \boldsymbol{\mu}^{j,l}||^2 +
    ||\sqrt{\mathbf{r}^{i,l}} - \sqrt{\mathbf{r}^{j,l}}||_2^2)^{1/2}.
\end{aligned}
    \label{eqa:dij}
\end{equation}

Large $d_{i,j}$ means the distribution distance between the $i-$th client and the $j$th client is large. Therefore, the larger $d_{i,j}$ is, the less similar the two clients are, which means the smaller $w_{i,j}$ is. And we set $\tilde{w}_{i,j}$ as the inverse of $d_{i,j}$, i.e. $\tilde{w}_{i,j} = 1 / d_{i,j}, j\neq i$. Normalize $\tilde{w}_i$ and we have 
\begin{equation}
\hat{w}_{i,j}= \frac{\tilde{w}_{i,j}}{\sum_{j=1,j\neq i}^N \tilde{w}_{i,j} },~\text{where} ~ j\neq i
\label{eqa:weight}
\end{equation}

For stability in training, we take $\boldsymbol{\psi}^{t*}$ into account for $\boldsymbol{\psi}^{t+1}$. 
We update $\boldsymbol{\psi}^{t+1}$ in a moving average style, and we set $w_{i,i} = \lambda$. Therefore,
\begin{equation}
    w_{i,j}= \begin{cases}
    \lambda,&i=j,\\
    (1 - \lambda) \times \hat{w}_{i,j}, & i\neq j.
    \end{cases} 
    \label{eqa:finalwei}
\end{equation}

We denote this weighting method as the original \emph{\method}.
Similarly, we can obtain the corresponding $\mathbf{W}$ using only the last layer $\mathbf{z}^i$ and we denote this variant as \emph{d-\method}.

\subsection{Discussion}
In some extreme cases, there may not exist a pre-trained model. In this situation, we can evaluate weights with models trained from several rounds of FedBN~\cite{li2021fedbn}.

\begin{figure}[htbp!]
	\centering
	\includegraphics[width=0.45\textwidth]{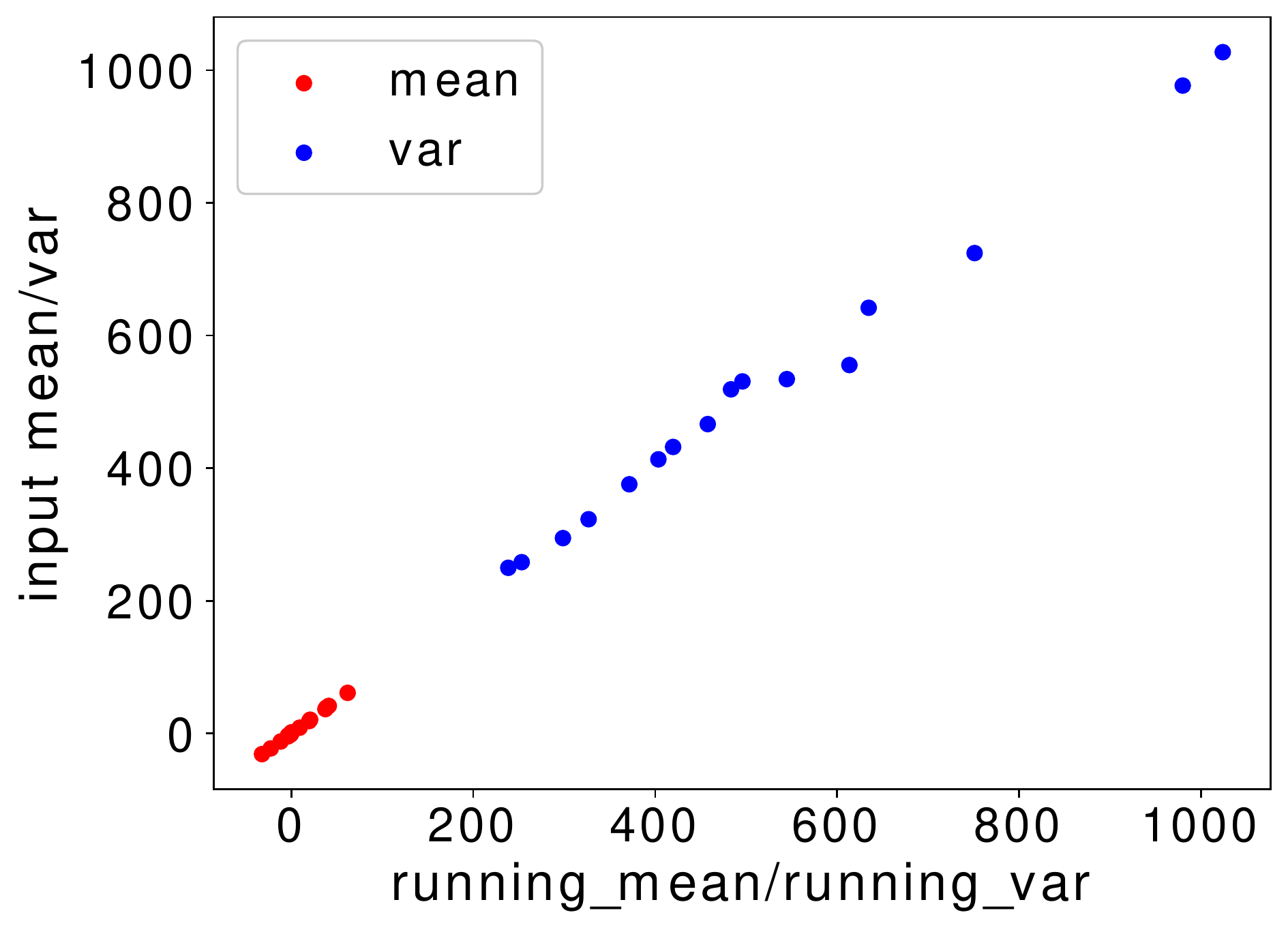}
	\caption{Running mean, running var of a BN layer and the inputs statistics of the corresponding layer in a client model.}
	\label{fig:fedbnweight}
\end{figure}

As we can see from \figurename~\ref{fig:fedbnweight}, the running mean of the BN layer has a positive correlation with the statistical mean of the corresponding layer's inputs. And the variance has a similar relationship. From this, we can use running means and running variances of the BN layers instead of the statistics respectively. Therefore, we can perform several rounds of FedBN~\cite{li2021fedbn} which preserves local batch normalization, and utilize parameters of BN layers to replace the statistics when there does not exist a pre-trained model. We denote this variant as \emph{f-\method}.

\section{Experiments}

We evaluate the performance of \method on five healthcare datasets in time series and image modalities\footnote{Code is released at \url{https://github.com/jindongwang/tlbook-code/tree/main/chap19_fl} and \url{https://github.com/microsoft/PersonalizedFL}.}. 
The statistical information of each dataset is shown in \tablename~\ref{tb-dataset}.

\begin{figure*}[t!]
	\centering
	\subfigure[OrganAMNIST]{
		\label{fig:madata}
		\includegraphics[width=0.3\textwidth]{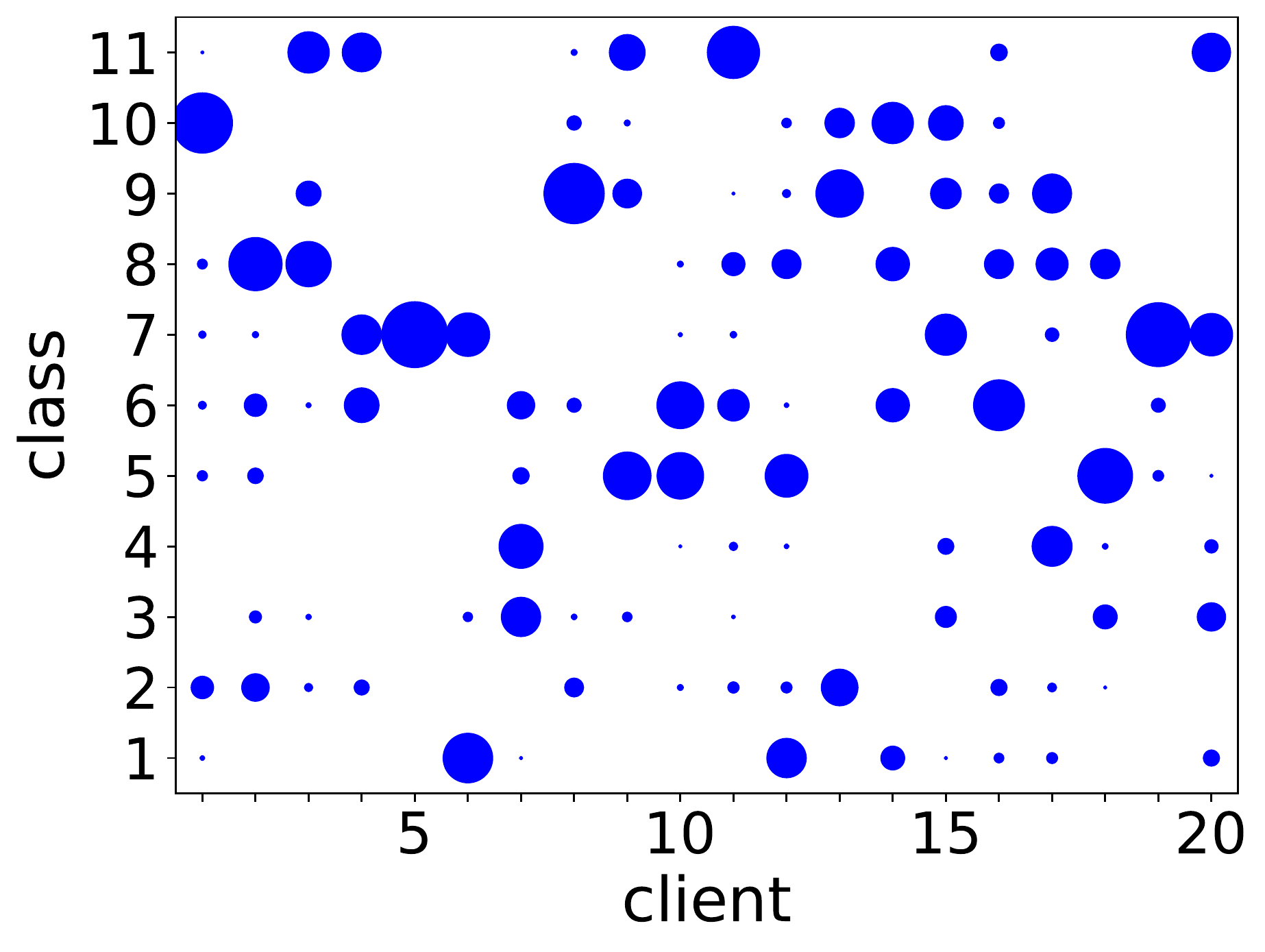}
	}
	\subfigure[OrganCMNIST]{
		\label{fig:mcdata}
		\includegraphics[width=0.3\textwidth]{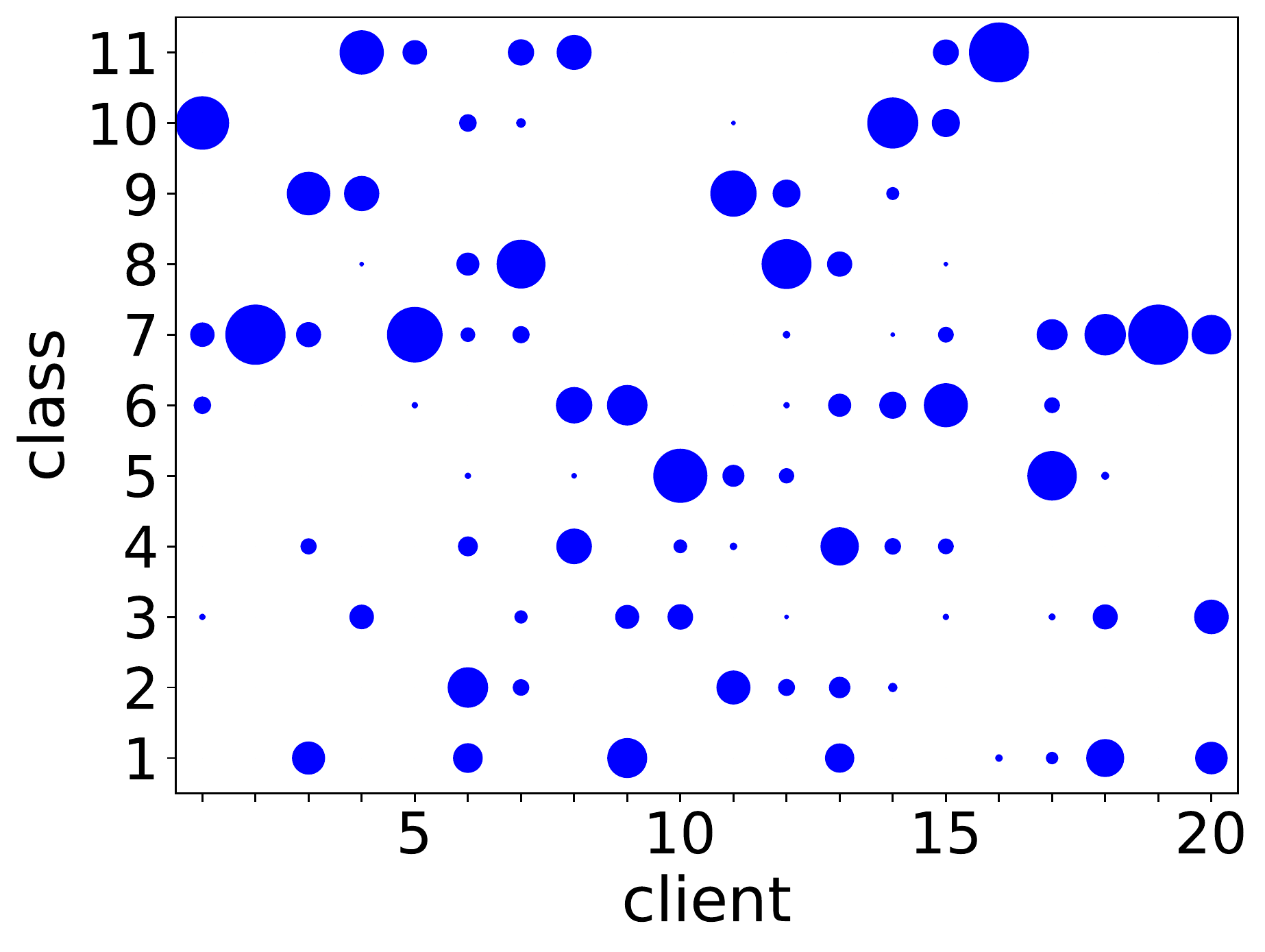}
	}
	\subfigure[OrganSMNIST]{
		\label{fig:msdata}
		\includegraphics[width=0.3\textwidth]{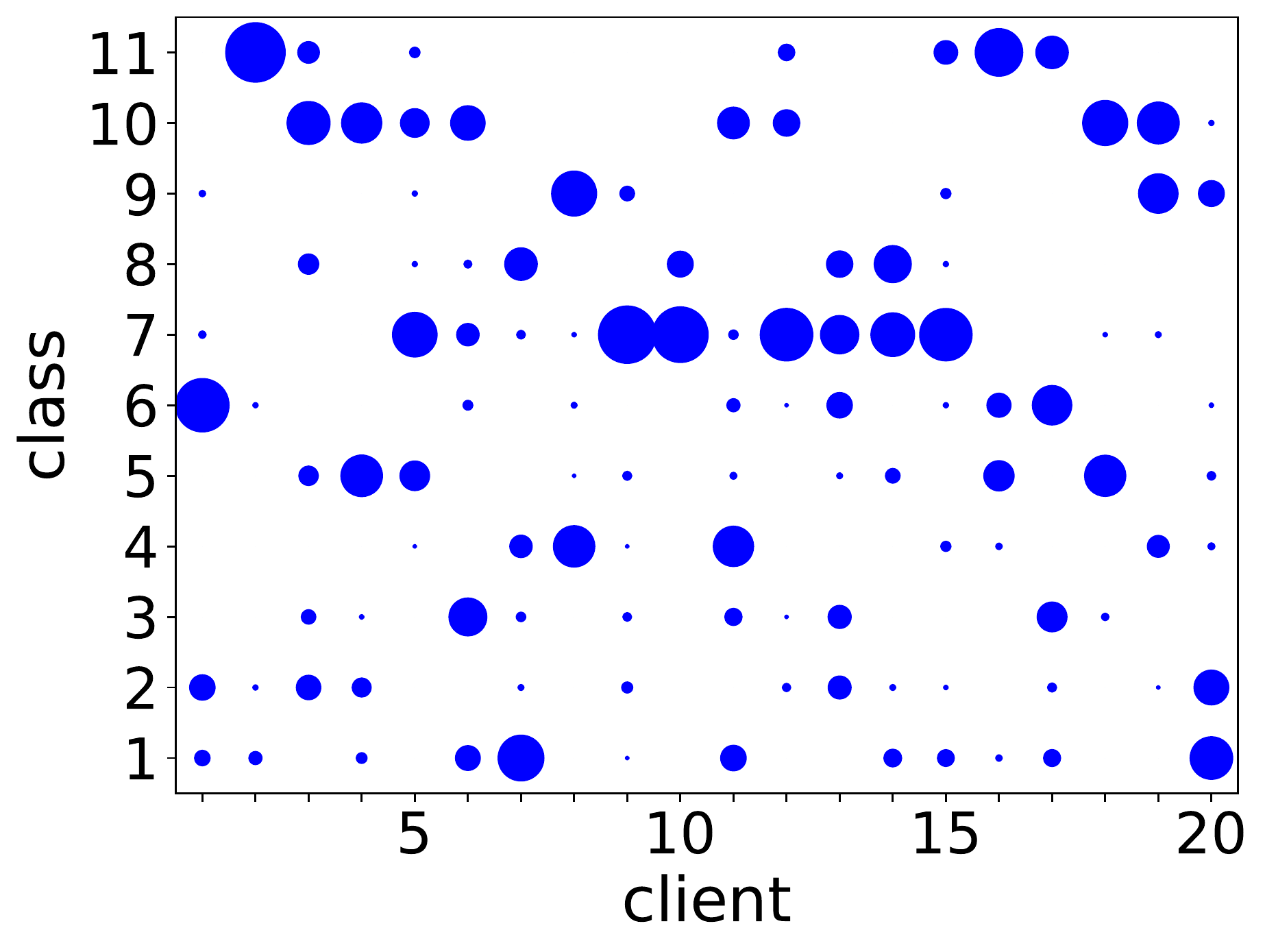}
	}
	
	\subfigure[PAMAP2]{
		\label{fig:pamapdata}
		\includegraphics[width=0.3\textwidth]{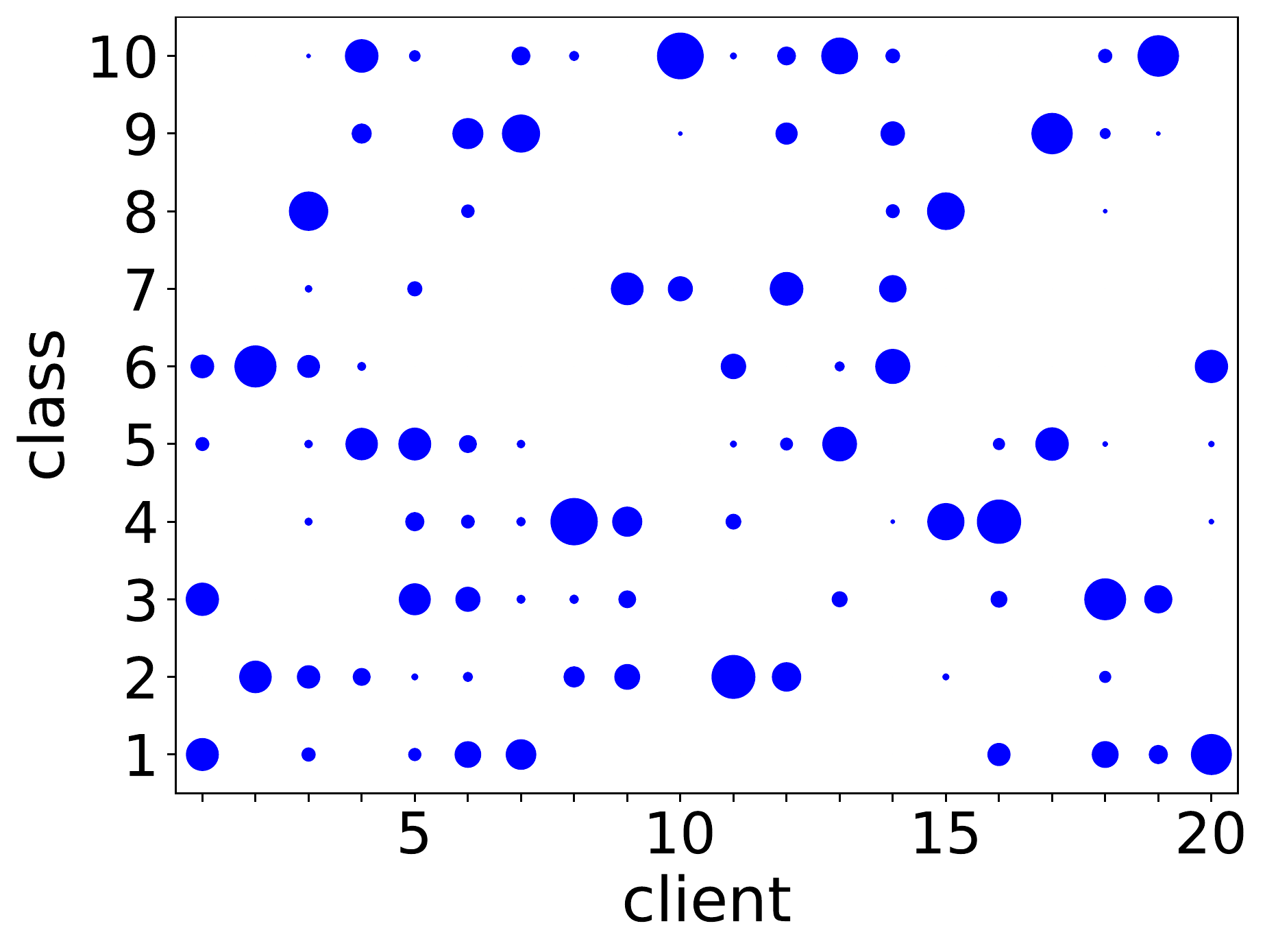}
	}
	\subfigure[COVID-19]{
		\label{fig:coviddata}
		\includegraphics[width=0.3\textwidth]{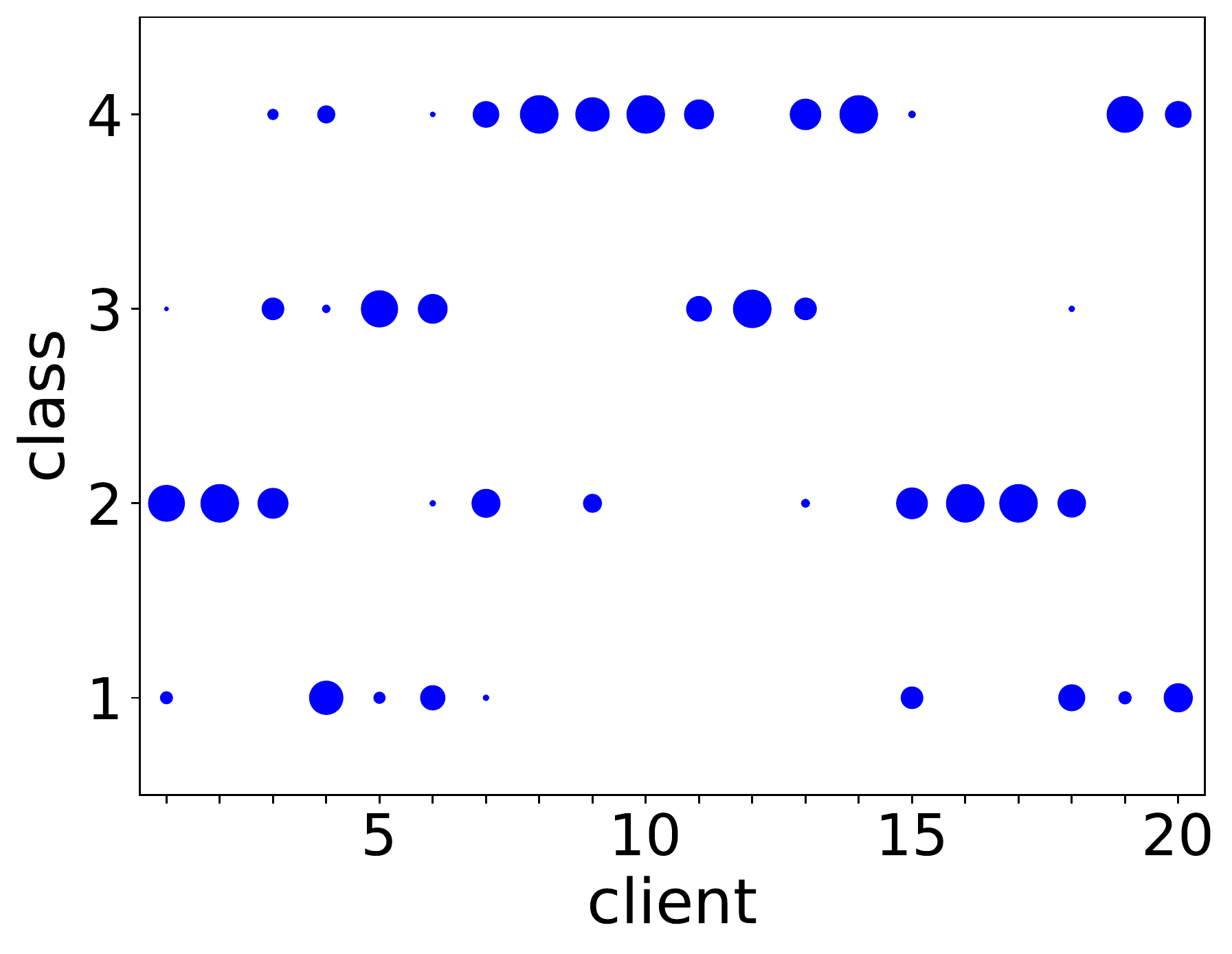}
	}
	\caption{The number of samples per class allocated to each client (indicated by dot size).}
	\label{fig:datasplit}
\end{figure*}

\subsection{Datasets}

\textbf{PAMAP2.} We adopt a public human activity recognition dataset called PAMAP2~\cite{reiss2012introducing}. The PAMAP2 dataset contains data of 18 different physical activities, performed by 9 subjects wearing 3 inertial measurement units and a heart rate monitor. We use data of 3 inertial measurement units which are collected at a constant rate of 100Hz to form data containing 27 channels. We exploit the sliding window technique and filter out 10 classes of data\footnote{We split PAMAP2 in this style mainly for two reasons. On the one hand, the data
numbers of the subjects are different which may introduce some other problems, e.g. some clients cannot be adequately evaluated. On the other hand, this splitting routing is widely adopted in much work~\cite{yurochkin2019bayesian,lin2021quasi}. We select 10 classes with the most samples.}. In order to construct the problem situation in \method, we use the Dirichlet distribution as in~\cite{yurochkin2019bayesian} to create disjoint non-iid splits. client training data. \figurename~\ref{fig:pamapdata} visualizes how samples are distributed among 20 clients. In each client, half of the data are used to train and the remaining data are for testing as in~\cite{li2021fedbn}.

\textbf{COVID-19.}
We also adopt a public COVID-19 posterior-anterior chest radiography images dataset~\cite{sait2020curated}. This is a combined curated dataset of COVID-19 Chest X-ray images obtained by collating 15 public datasets and it contains 9,208 instances of four classes (1,281 COVID-19 X-Rays, 3,270 Normal X-Rays, 1,656 viral-pneumonia X-Rays, and 3,001 bacterial-pneumonia X-Rays) in total. In order to construct the problem situation in \method, we split the dataset similar to PAMAP2. \figurename~\ref{fig:coviddata} visualizes how samples are distributed among 20 clients for COVID-19. Note that this dataset is more unbalanced in classes which is an ideal testbed to test the performance under label shift (i.e., imbalanced class distribution for different clients). In each client, half of the data are used to train and the remaining data are for testing.

\textbf{MedMnist.} 
MedMNIST~\cite{medmnistv1,medmnistv2} is a large-scale MNIST-like collection of standardized biomedical images, including 12 datasets for 2D and 6 datasets for 3D. All images are $28\times 28$ (2D) or $28\times 28 \times 28$ (3D). We choose 3 datasets which have most classes from 12 2D datasets: OrganAMNIST, OrganCMNIST, OrganSMNIST~\cite{bilic2019liver,xu2019efficient}. These three datasets are all about Abdominal CT images and all contain 11 classes. There are 58,850, 23,660 and 25,221 samples respectively. As operations in PAMAP2, each dataset is split into 20 clients with Dirichlet distributions, and \figurename~\ref{fig:madata}-\ref{fig:msdata} visualizes how samples are distributed for OrganAMNIST, OrganCMNIST, and OrganSMNIST respectively. In each client, half of the data are used to train and the remaining data are for testing.

\begin{table}[t!]
\centering
\caption{Statistical information of five datasets.}
\resizebox{.5\textwidth}{!}{%
\begin{tabular}{ccrr}
\toprule
Dataset & Type & \#Class & \#Sample  \\ \midrule
PAMAP2    & Sensor-based time series & 18           & 3,850,505  \\
COVID-19   & Image         & 4         & 9,208 \\
OrganAMNIST  & Image         & 11        & 58,850 \\
OrganCMNIST  & Image         & 11         & 23,660 \\
OrganSMNIST  & Image         & 11         & 25,221 \\
\bottomrule
\end{tabular}%
}
\label{tb-dataset}
\end{table}

\begin{table}[!t]
\centering
\caption{Activity recognition results on PAMAP2. Bold and underline mean the best and
second-best results, respectively.}
\label{tab:my-table-pamap}
\begin{tabular}{ccccccc}
\toprule
Client & Base           & FedAvg & FedBN       & FedProx     & FedPer         & \method         \\ \midrule
1      & \textbf{92.86} & 60.27  & 60.72       & 60.5        & 48.31          & \underline{77.2}     \\
2      & 17.68          & 62.36  & 62.59       & 62.36       & \textbf{97.51} & \underline{77.55}    \\
3      & \textbf{100}   & 50.56  & 50.34       & 50.34       & 61.4           & \underline{77.43}    \\
4      & \textbf{83.52} & 73.98  & 73.53       & 73.98       & 47.29          & \underline{79.64}    \\
5      & 18.78          & 74.27  & \underline{74.72} & 73.81       & 58.47          & \textbf{81.94} \\
6      & \underline{77.66}    & 62.9   & 62.44       & 61.76       & 23.98          & \textbf{79.86} \\
7      & \textbf{95.05} & 64.03  & 62.9        & 63.57       & 49.55          & \underline{86.2}     \\
8      & 17.58          & 87.78  & 88.24       & 87.78       & \underline{91.86}    & \textbf{95.02} \\
9      & \textbf{92.39} & 74.49  & 74.27       & 74.27       & 51.24          & \underline{85.33}    \\
10     & \textbf{93.37} & 64.71  & 64.48       & 64.71       & \underline{77.6}     & 69.23          \\
11     & 29.12          & 65.24  & 65.69       & 66.37       & \underline{89.16}    & \textbf{91.42} \\
12     & \textbf{84.78} & 63.35  & 62.9        & 63.12       & 57.92          & \underline{79.41}    \\
13     & \textbf{98.9}  & 68.33  & 68.33       & 68.33       & 42.53          & \underline{74.43}    \\
14     & 24.18          & 64.79  & 65.24       & \underline{65.69} & 49.44          & \textbf{69.75} \\
15     & \textbf{98.91} & 63.12  & 62.44       & 62.44       & 58.6           & \underline{81.67}    \\
16     & \textbf{98.9}  & 85.26  & 85.94       & 85.49       & 86.62          & \underline{94.1}     \\
17     & 41.44          & 66.21  & 65.99       & 66.21       & \underline{77.32}    & \textbf{82.77} \\
18     & \textbf{93.62} & 59.64  & 59.64       & 59.41       & 52.38          & \underline{75.74}    \\
19     & \textbf{85.71} & 67.87  & 68.1        & 67.87       & 73.08          & \underline{77.15}    \\
20     & 37.02          & 72.46  & 72.69       & 72.46       & \textbf{97.52} & \underline{86.91}    \\
avg    & \underline{69.07}    & 67.58  & 67.56       & 67.52       & 64.59          & \textbf{81.14}\\
\bottomrule
\end{tabular}
\end{table}

\begin{table}[!t]
\centering
\caption{Accuracy on OrganAMNIST. Bold and underline mean the best and
second-best results, respectively.}
\label{tab:my-table-OrganA}
\begin{tabular}{ccccccc}
\toprule
Client & Base  & FedAvg      & FedBN          & FedProx     & FedPer       & \method         \\ \midrule
1      & 48.35 & 80.03       & \textbf{96.06} & 80.37       & \underline{83.22}  & 81.86          \\
2      & 55.25 & 92.46       & \underline{93.14}    & 92.26       & 78.68        & \textbf{94.5}  \\
3      & 34.04 & 86.15       & \underline{96.27}    & 86.22       & 71.08        & \textbf{97.08} \\
4      & 61.54 & 77.65       & \underline{87.91}    & 77.58       & 41.71        & \textbf{88.52} \\
5      & 41.44 & 92.32       & \textbf{100}   & 92.66       & \textbf{100} & \underline{99.93}    \\
6      & 52.13 & 84.38       & \textbf{97.55} & 84.92       & 80.57        & \underline{95.52}    \\
7      & 42.31 & \underline{83.42} & 50.07          & 82.95       & 64.06        & \textbf{87.84} \\
8      & 48.9  & 92.81       & \textbf{97.15} & 92.94       & 87.86        & \underline{96.95}    \\
9      & 38.04 & 74.66       & \underline{84.1}     & 74.39       & 62.16        & \textbf{85.46} \\
10     & 38.12 & 72.27       & \underline{82.98}    & 72.34       & 53.22        & \textbf{87.53} \\
11     & 59.89 & 74.88       & \underline{90.36}    & 74.88       & 66.33        & \textbf{91.79} \\
12     & 59.78 & 78.41       & \textbf{90.56} & 78.28       & 72.84        & \underline{89.75}    \\
13     & 44.75 & 91.17       & \textbf{97.69} & 91.04       & 78.14        & \underline{97.42}    \\
14     & 52.2  & 83.76       & \textbf{92.26} & 83.9        & 61.28        & \underline{90.29}    \\
15     & 53.26 & 89.61       & 87.84          & \underline{90.42} & 58.29        & \textbf{94.29} \\
16     & 70.88 & 83.31       & \textbf{93.62} & 83.18       & 73.54        & \underline{91.59}    \\
17     & 36.46 & \underline{92.93} & 62.77          & \underline{92.93} & 45.92        & \textbf{94.97} \\
18     & 46.81 & 77.99       & \textbf{96.94} & 77.92       & 78.12        & \underline{96.67}    \\
19     & 31.32 & 92.05       & \textbf{96.94} & 92.26       & 93.95        & \underline{96.47}    \\
20     & 45.3  & 81.02       & \underline{91.32}    & 80.75       & 49.36        & \textbf{93.97} \\
avg    & 48.04 & 84.06       & \underline{89.28}    & 84.11       & 70.02        & \textbf{92.62}\\ \bottomrule
\end{tabular}
\end{table}

\begin{table}[!t]
\centering
\caption{Accuracy on OrganCMNIST. Bold and underline mean the best and
second-best results, respectively.}
\label{tab:my-table-OrganC}
\begin{tabular}{ccccccc}\toprule
Client & Base        & FedAvg      & FedBN          & FedProx     & FedPer         & \method         \\ \midrule
1      & 32.61       & 79.22       & \underline{90.54}    & 79.73       & 77.87          & \textbf{94.59} \\
2      & 52.17       & 95.61       & \textbf{100}   & \underline{95.95} & \textbf{100}   & \textbf{100}   \\
3      & 47.1        & 85.83       & \underline{88.7}     & 85.83       & 72.85          & \textbf{93.93} \\
4      & 37.23       & 84.34       & \textbf{96.97} & 84.01       & 74.41          & \underline{96.3}     \\
5      & 48.91       & 92.41       & \underline{96.46}    & 92.24       & 84.49          & \textbf{97.13} \\
6      & 51.45       & 65.6        & \underline{75.89}    & 65.6        & 53.96          & \textbf{85.5}  \\
7      & 64.49       & 86.22       & \underline{86.72}    & 86.55       & 76.64          & \textbf{89.08} \\
8      & 45.65       & 50.93       & \underline{61.38}    & 50.59       & 38.45          & \textbf{92.07} \\
9      & 26.09       & 81.28       & \textbf{91.23} & 80.61       & 45.03          & \underline{86.68}    \\
10     & 57.25       & 54.56       & \textbf{93.41} & 54.9        & 80.07          & \underline{91.89}    \\
11     & 54.89       & 79.73       & \textbf{94.76} & 79.73       & 83.78          & \underline{91.39}    \\
12     & 50.72       & \underline{90.56} & \textbf{95.78} & 90.39       & 67.28          & 89.38          \\
13     & \underline{66.67} & 49.33       & 54.71          & 49.16       & 52.36          & \textbf{89.73} \\
14     & 37.16       & 74.32       & \textbf{88.34} & 73.99       & 70.61          & \underline{87.33}    \\
15     & 58.57       & 75.93       & \underline{86.36}    & 75.59       & 51.52          & \textbf{89.9}  \\
16     & 52.9        & 81.25       & \underline{98.82}    & 80.91       & \textbf{98.99} & \underline{98.82}    \\
17     & 50          & 67.45       & \textbf{83.81} & 67.45       & 70.66          & \underline{81.11}    \\
18     & 52.9        & 88.18       & \underline{90.71}    & 88.18       & 65.37          & \textbf{91.89} \\
19     & 26.28       & 94.26       & \textbf{100}   & 94.26       & \textbf{100}   & \underline{99.83}    \\
20     & 59.78       & \underline{90.25} & 88.91          & \underline{90.25} & 58.49          & \textbf{93.78} \\
avg    & 48.64       & 78.36       & \underline{88.18}    & 78.3        & 71.14          & \textbf{92.02} \\ \bottomrule
\end{tabular}
\end{table}

\begin{table}[!t]
\centering
\caption{Accuracy on OrganSMNIST. Bold and underline mean the best and
second-best results, respectively.}
\label{tab:my-table-OrganS}
\begin{tabular}{ccccccc}
\toprule
Client & Base  & FedAvg      & FedBN          & FedProx     & FedPer         & \method         \\ \midrule
1      & 25.27 & 47.39       & \textbf{91.76} & 39.94       & 85.74          & \underline{90.65}    \\
2      & 32.04 & 55.7        & \textbf{96.04} & 62.82       & \underline{94.78}    & 93.51          \\
3      & 30.85 & 63.13       & \underline{71.84}    & 66.3        & 52.69          & \textbf{73.89} \\
4      & 41.21 & 59.65       & \textbf{78.8}  & 61.71       & 64.56          & \underline{75.32}    \\
5      & 37.02 & 68.45       & \underline{80.28}    & 73.03       & 68.61          & \textbf{83.12} \\
6      & 40.43 & 48.02       & \underline{77.02}    & 52.3        & 58             & \textbf{80.67} \\
7      & 38.46 & 78.2        & \underline{83.57}    & 71.25       & 78.2           & \textbf{85.47} \\
8      & 40.11 & 57.44       & 53.48          & 41.93       & \textbf{96.52} & \underline{96.2}     \\
9      & 42.39 & 83.73       & \textbf{94.94} & 87.99       & \underline{92.42}    & \textbf{94.94} \\
10     & 24.86 & 88.45       & \underline{99.21}    & 87.97       & 97.47          & \textbf{99.37} \\
11     & 38.46 & \underline{60.79} & 45.04          & 37.17       & 54.49          & \textbf{76.06} \\
12     & 41.3  & 79.15       & 81.67    & \underline{83.57}       & 78.04          & \textbf{88.47} \\
13     & 43.65 & 66.98       & \underline{79.94}    & 67.93       & 58.29          & \textbf{80.57} \\
14     & 44.51 & 84.99       & \textbf{95.73} & 86.57       & 84.52          & \underline{92.58}    \\
15     & 44.57 & 80.7        & \textbf{88.77} & 84.49       & 73.1           & \underline{87.34}    \\
16     & 48.9  & 52.06       & \textbf{73.58} & 53.64       & \underline{71.36}    & \textbf{73.58} \\
17     & 36.46 & 36.55       & \textbf{78.48} & 39.56       & 61.55          & \underline{76.42}    \\
18     & 45.74 & 58.7        & \textbf{79.43} & 60.44       & \underline{69.3}     & \textbf{79.59} \\
19     & 30.77 & 53.48       & \underline{86.23}    & 61.87       & 64.72          & \textbf{91.46} \\
20     & 35.91 & 72.51       & 72.99          & \underline{77.57} & \textbf{86.57} & 68.4           \\
avg    & 38.15 & 64.8        & \underline{80.44}    & 64.9        & 74.55          & \textbf{84.38}\\ \bottomrule
\end{tabular}
\end{table}

\subsection{Implementations Details and Comparison Methods}
For PAMAP2, we adopt a CNN for training and predicting. The network is composed of two convolutional layers, two pooling layers, two batch normalization layers, and two fully connected layers. For three MedMNIST datasets, we all adopt LeNet5~\cite{lecun1998gradient}. For COVID-19, we adopt Alexnet~\cite{krizhevsky2012imagenet}. We use a three-layer fully connected neural network as the classifier with two BN layers after the first two fully connected layers following~\cite{li2021fedbn}. 
For model training, we use the cross-entropy loss and SGD optimizer with a learning rate of $10^{-2}$. 
If not specified, our default setting for local update epochs is $E = 1$ where $E$ means training epochs in one round. 
And we set $\lambda=0.5$ for our method, since we can see that $\lambda$ has few influences on accuracy and it only affects convergence speeds in the appendix.
In addition, we randomly select $20\%$ of the data to train a model of the same architecture as the pre-trained model.
We run three trials to record the average results.

We compare three extensions of our method with five methods including common FL methods and some FL methods designed for non-iid data:
\begin{itemize}
    \item Base: Each client uses local data to train its local models without federated learning.
    \item FedAvg~\cite{mcmahan2017communication}: The server aggregates all client models without any particular operations for non-iid data.
    \item FedProx~\cite{li2018federated}: Allow partial information aggregation and add a proximal term to FedAvg.
    \item FedPer~\cite{arivazhagan2019federated}: Each client preserves some local layers.
    \item FedBN~\cite{li2021fedbn}: Each client preserves the local batch normalization.
\end{itemize}

\begin{figure}[t!]
	\centering
	\includegraphics[width=0.45\textwidth]{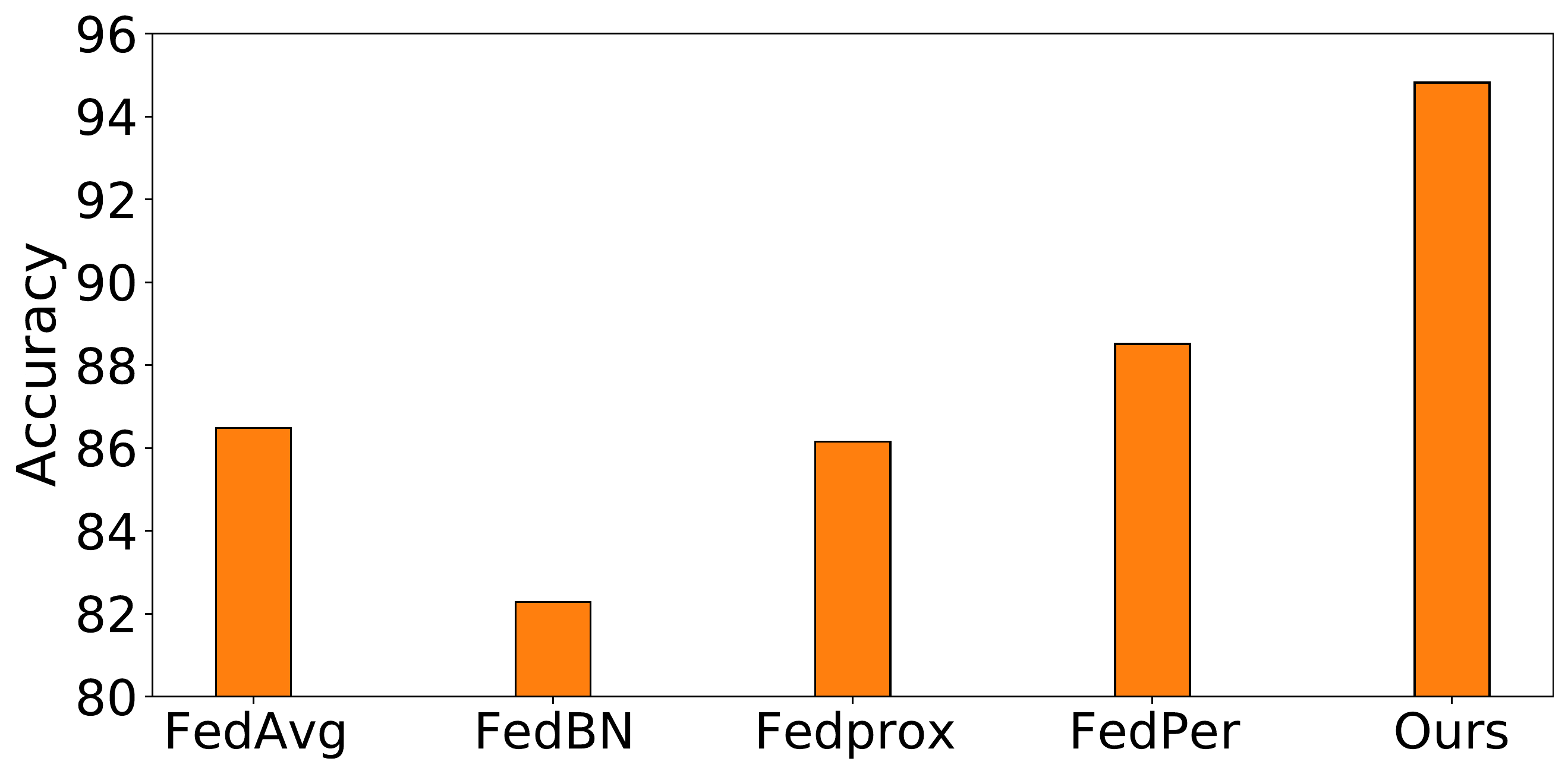}
	\caption{Average accuracy of 20 clients on COVID-19.}
	\label{fig:covidacc}
\end{figure}

\begin{figure*}[t!]
	\centering
	\subfigure[The weighting technique]{
		\label{fig:weffa}
		\includegraphics[height=0.16\textwidth]{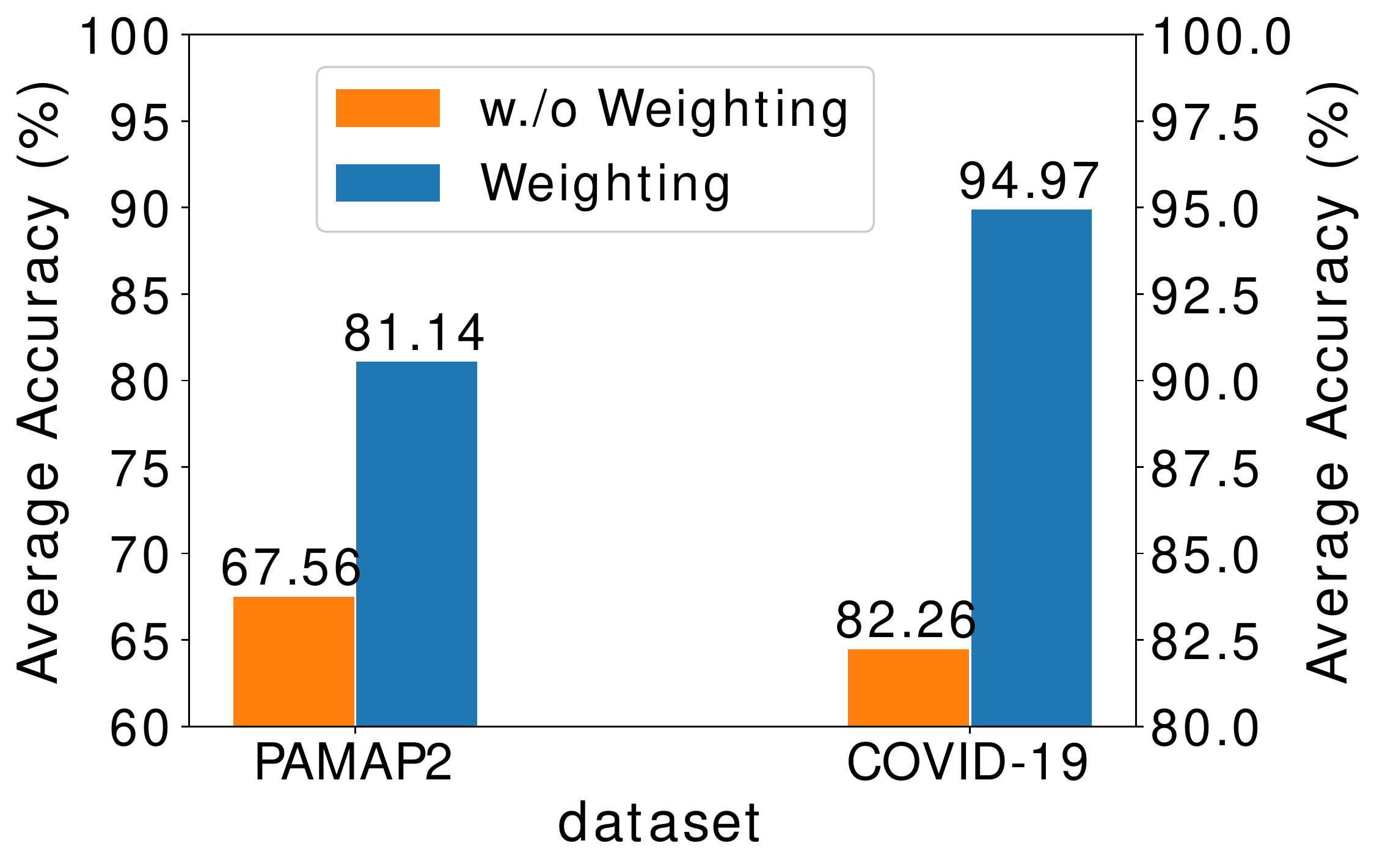}
	}
	\subfigure[Different clients]{
		\label{fig:weff}
		\includegraphics[height=0.16\textwidth]{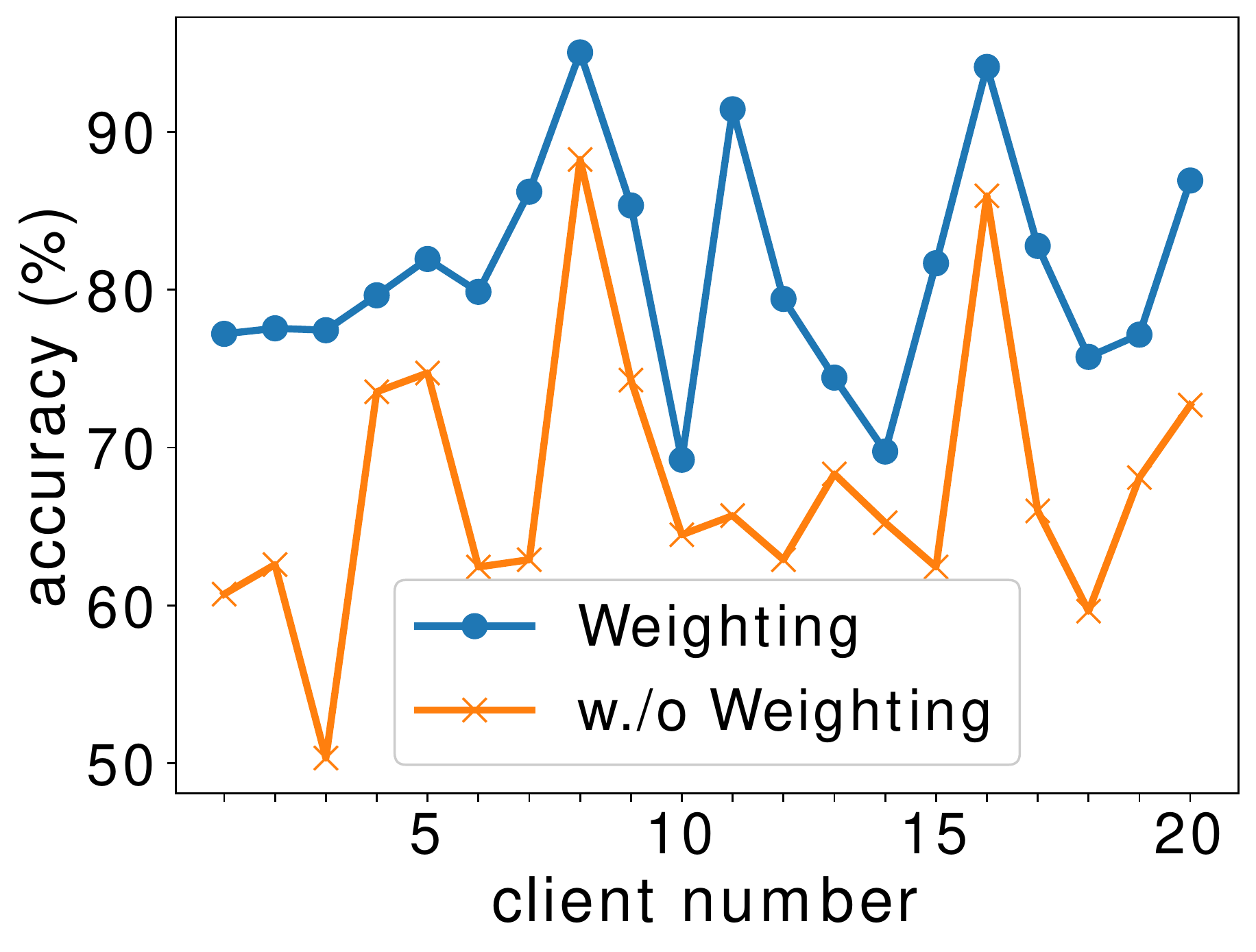}
	}
	\subfigure[The local BN sharing]{
		\label{fig:bneffa}
		\includegraphics[height=0.16\textwidth]{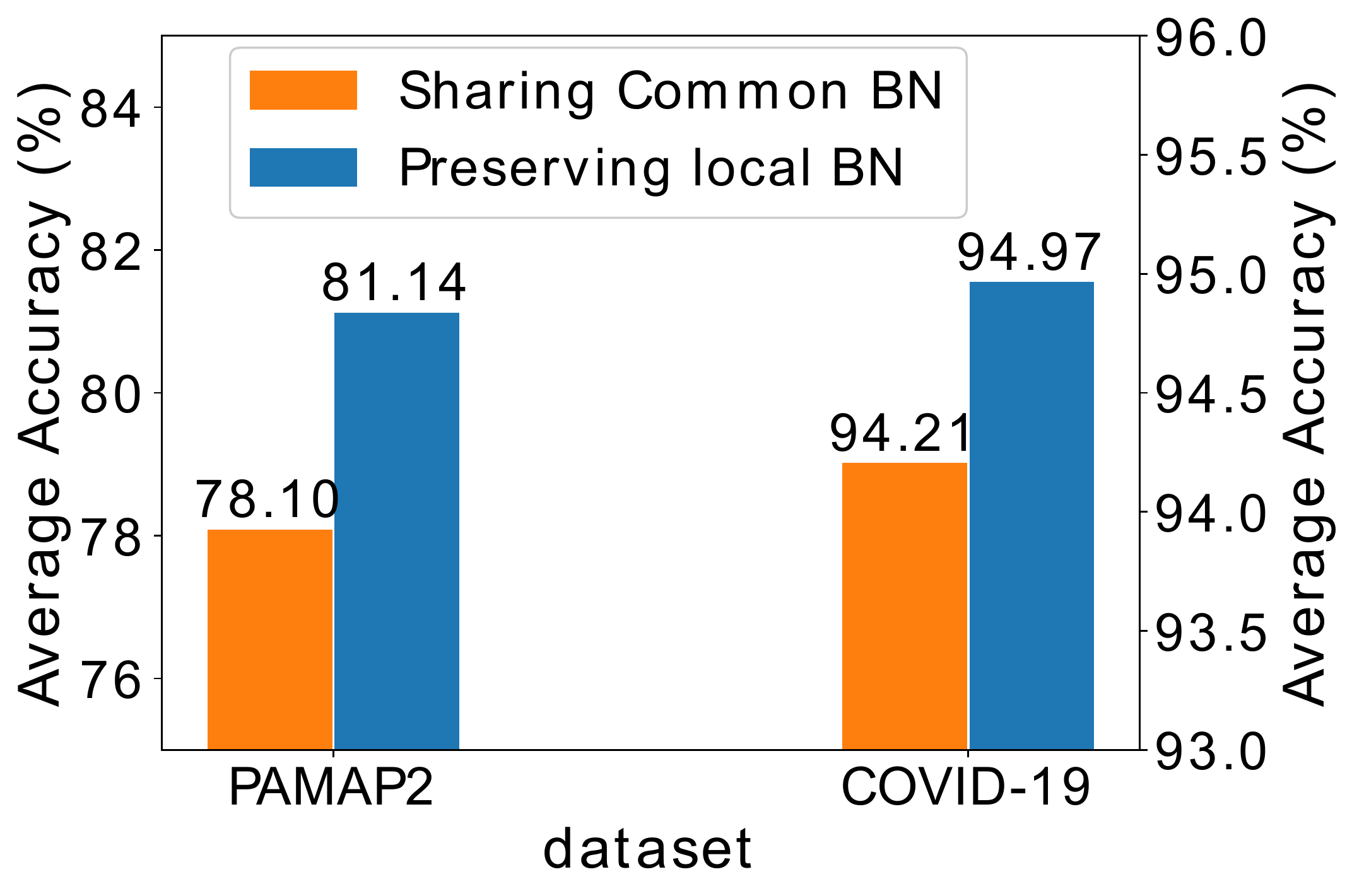}
	}
	\subfigure[Client Acc on PAMAP2]{
		\label{fig:bneff}
		\includegraphics[height=0.16\textwidth]{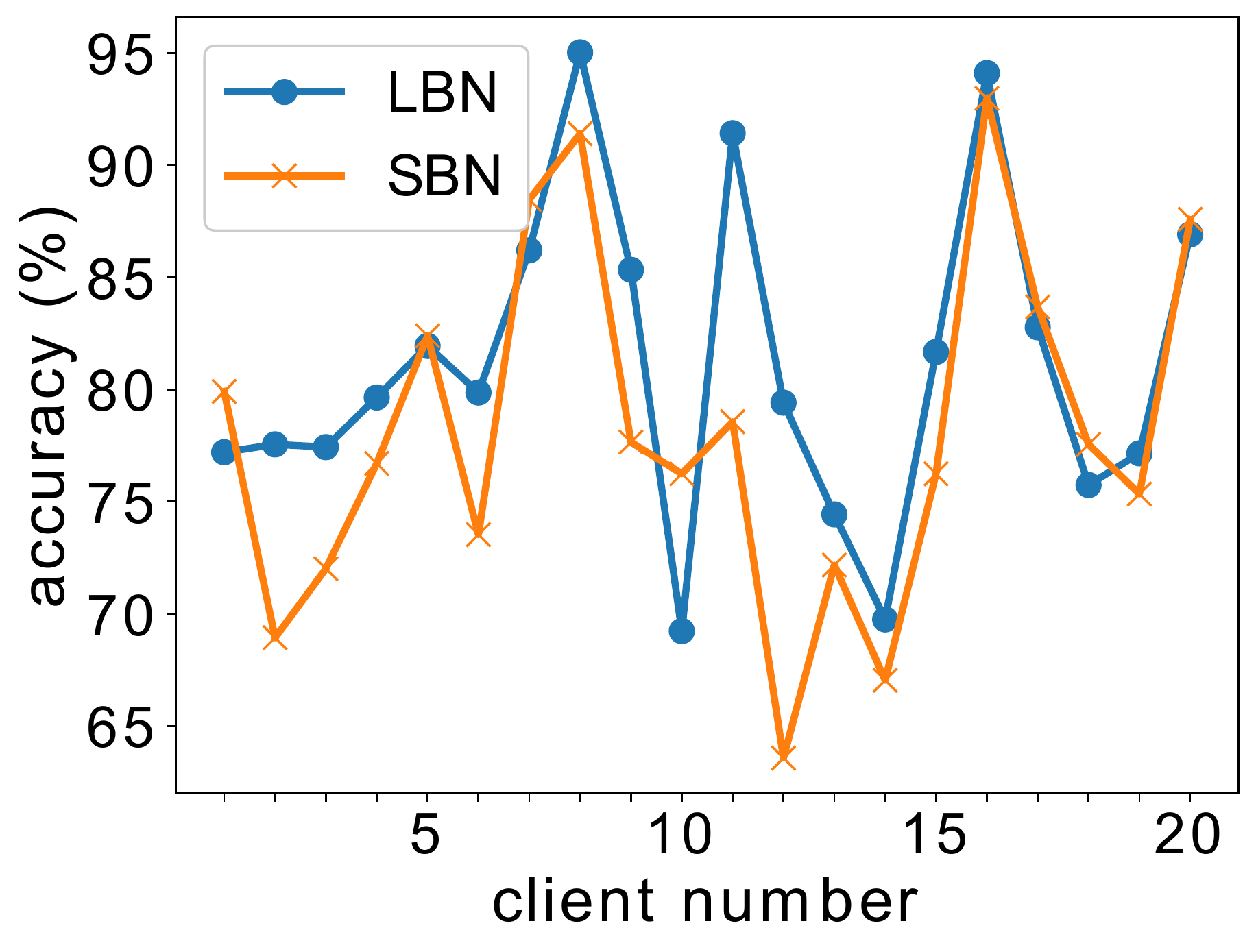}
	}
	\caption{The effects of weighting and preserving local batch normalization. Each point has equal status in \figurename~\ref{fig:weff} and \figurename~\ref{fig:bneff}. We use a line chart just for better visualization effects but not trends.}
	\label{fig:ablat1}
\end{figure*}

\begin{figure}[t!]
	\centering
	\subfigure[Dataset Split]{
		\label{fig:split}
		\includegraphics[height=0.16\textwidth]{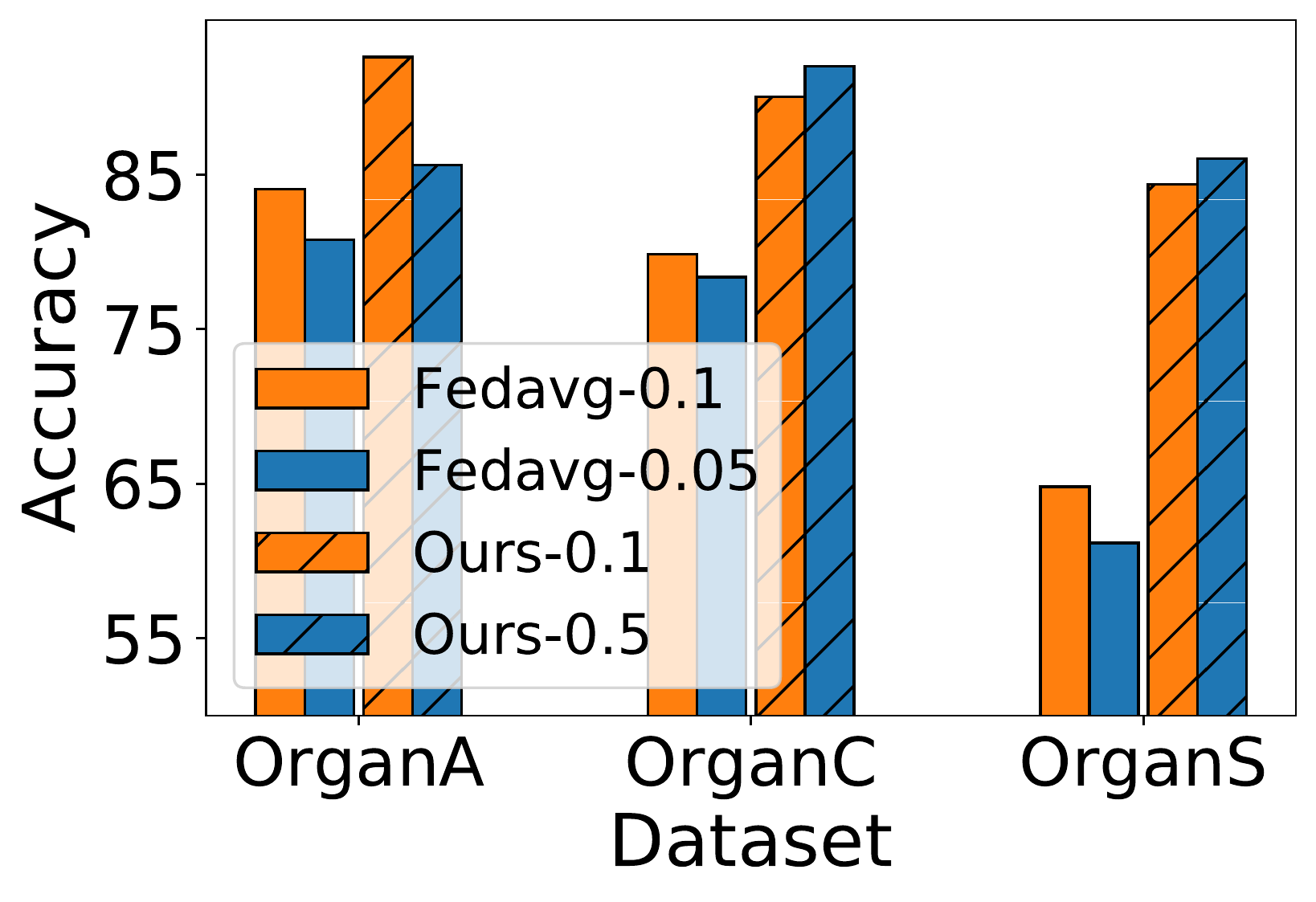}
	}
	\subfigure[Iteration-Round]{
		\label{fig:comcost}
		\includegraphics[height=0.16\textwidth]{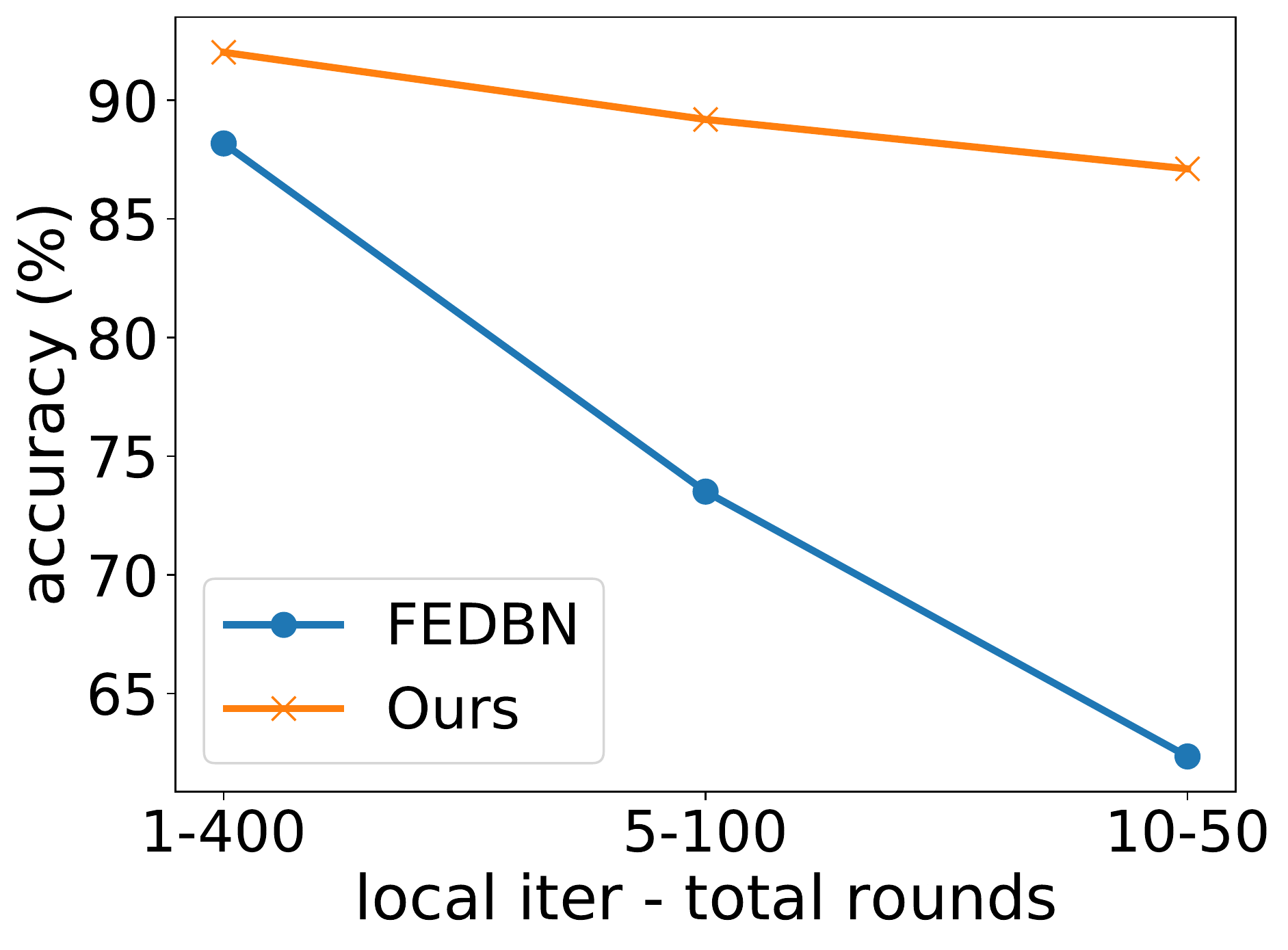}
	}
	\caption{Influence of Dataset split and Iteration-Round.}
	\label{fig:inf-dis}
\end{figure}

\subsection{Classification Accuracy}

The classification results for each client on PAMAP2 are shown in \tablename~\ref{tab:my-table-pamap}. From these results, we have the following observations: 1) Our method achieves the best results on average. It is obvious that our method significantly outperforms other methods with a remarkable improvement (over $\mathbf{10}\%$ on average). 2) In some clients, the base method achieves the best test accuracy. As it can be seen from \figurename~\ref{fig:pamapdata}, the distributions on the clients are very inconsistent, which inevitably leads to the various difficulty levels in different clients. And some distributions in the corresponding clients are so easy that only utilizing the local data can achieve the ideal effects. 3) FedBN does not achieve the desired results. This could be caused by that FedBN is designed for the feature shifts while our experiments are mainly set in the label shifts. 

The classification results for each client on three MedMNIST datasets are shown in \tablename~\ref{tab:my-table-OrganA}, \ref{tab:my-table-OrganC}, \ref{tab:my-table-OrganS} respevtively. From these results, we have the following observations: 1) Our method significantly outperforms other methods with a remarkable improvement (over $\mathbf{3.5}\%$ on average). 2) For all these three benchmarks, Base achieves the worst average accuracy, which demonstrates Base without communicating with each other does not have enough information for these relatively difficult tasks. 3) FedBN achieves the second best results on all three benchmarks. This could be because that there exist feature shifts among clients.


The classification results for each client on COVID-19 are shown in \figurename~\ref{fig:covidacc}. From these results, we have the following observations: 1) Our method achieves the best average accuracy which outperforms the second-best method FedPer by $\mathbf{6.3}\%$ on average accuracy. 2) FedBN gets the worst results. This demonstrates that FedBN is not good at dealing with label shifts where label distributions of each client are different, which is a challenging situation. FedBN does not consider the similarities among different clients. From \figurename~\ref{fig:coviddata}, we can see that label shifts are serious in COVID-19 since it only has four classes.

\subsection{Analysis and discussion}

We consider the influence of data splits and local iterations in this section. As shown in \figurename~\ref{fig:split}, we evaluate Fedavg and \method on three MedMNIST benchmarks with two different splits: $\alpha=0.1$ and $\alpha=0.05$ respectively. Smaller $\alpha$ means distributions among clients are more different from each other. \figurename~\ref{fig:split} demonstrates that the performance of Fedavg which does not consider data non-iid will drop when encountering clients with greater different distributions while our method is not affected much by the degree of data non-iid, which means our method may be more robust. \figurename~\ref{fig:comcost} shows the influence of local iterations and total rounds on FedBN and our method. It is obvious that FedBN drops seriously with more local iterations and fewer communication rounds while our method declines slowly, which means when limiting communication costs, our method may be more effective. 

\begin{figure}[t!]
	\centering
	\subfigure[PAMAP2]{
		\label{fig:p-other}
		\includegraphics[height=0.2\textwidth]{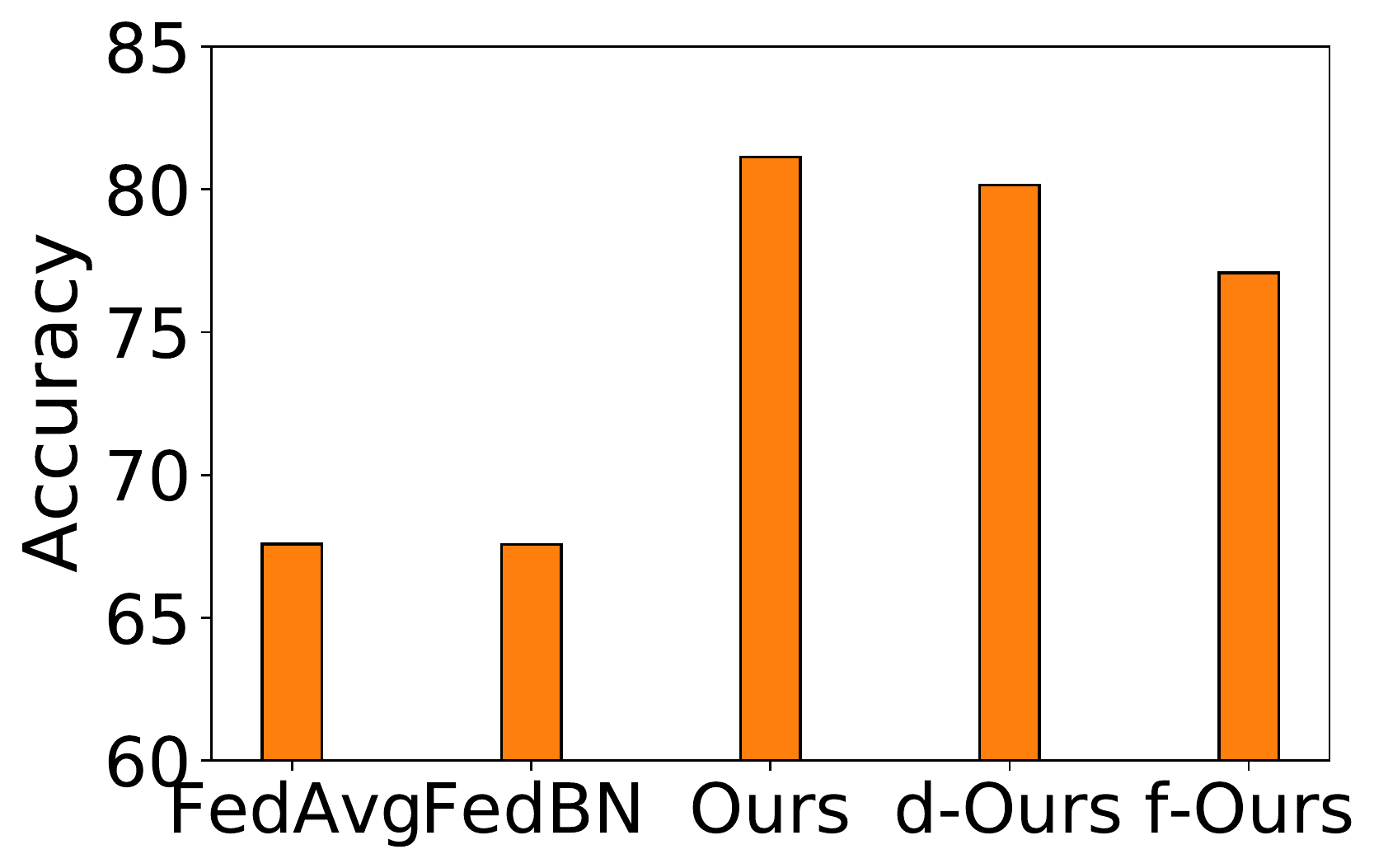}
	}
	\subfigure[COVID]{
		\label{fig:c-other}
		\includegraphics[height=0.2\textwidth]{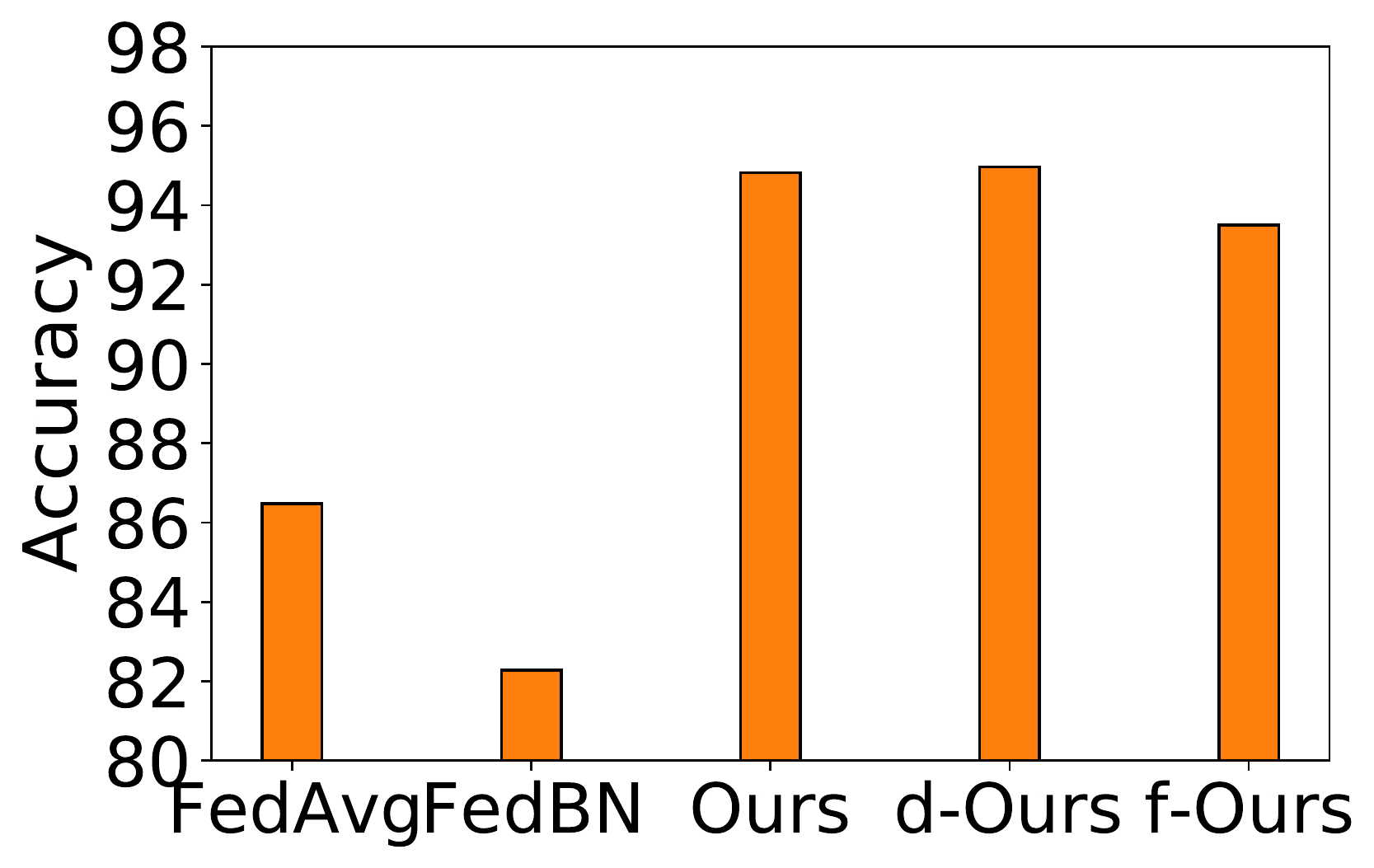}
	}
	\caption{Evaluating two variants of \method.}
	\label{fig:other weighting}
\end{figure}

\begin{figure*}[t!]
	\centering
	\subfigure[Client Number]{
		\label{fig:s-clients}
		\includegraphics[width=0.23\textwidth]{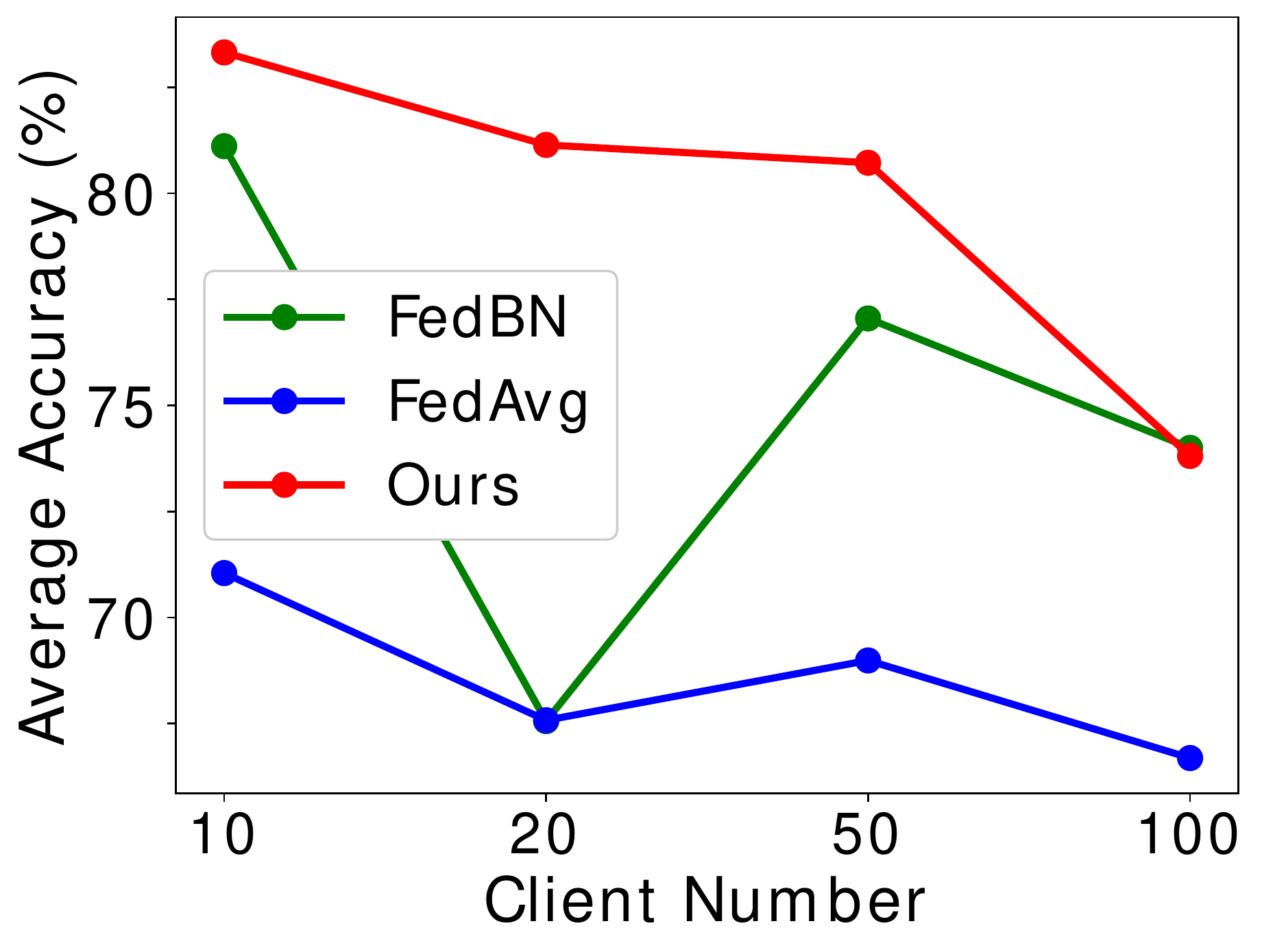}
	}
	\subfigure[Local Epochs]{
		\label{fig:s-lepc}
		\includegraphics[width=0.23\textwidth]{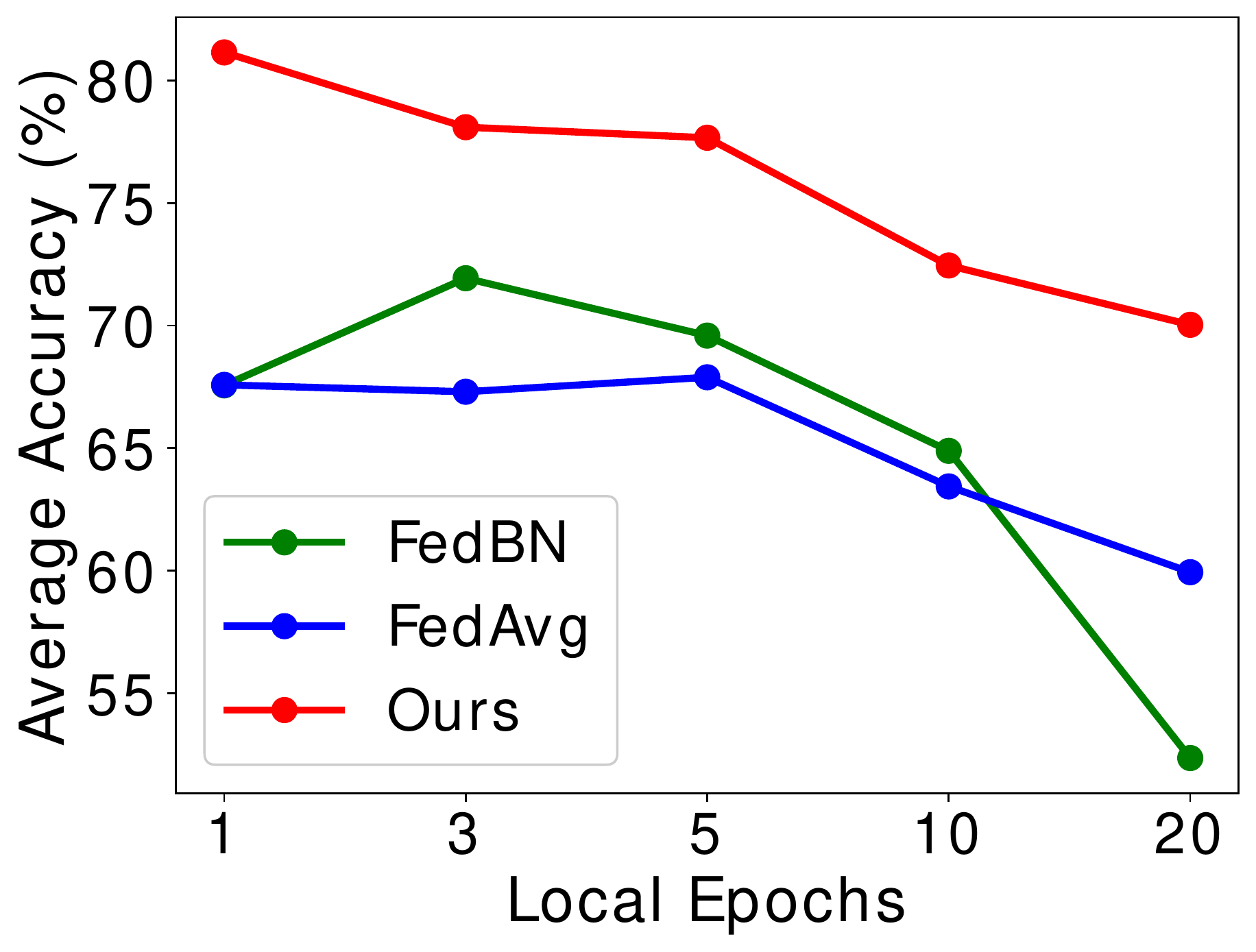}
	}
	\subfigure[$\lambda$]{
		\label{fig:s-lambda}
		\includegraphics[width=0.23\textwidth]{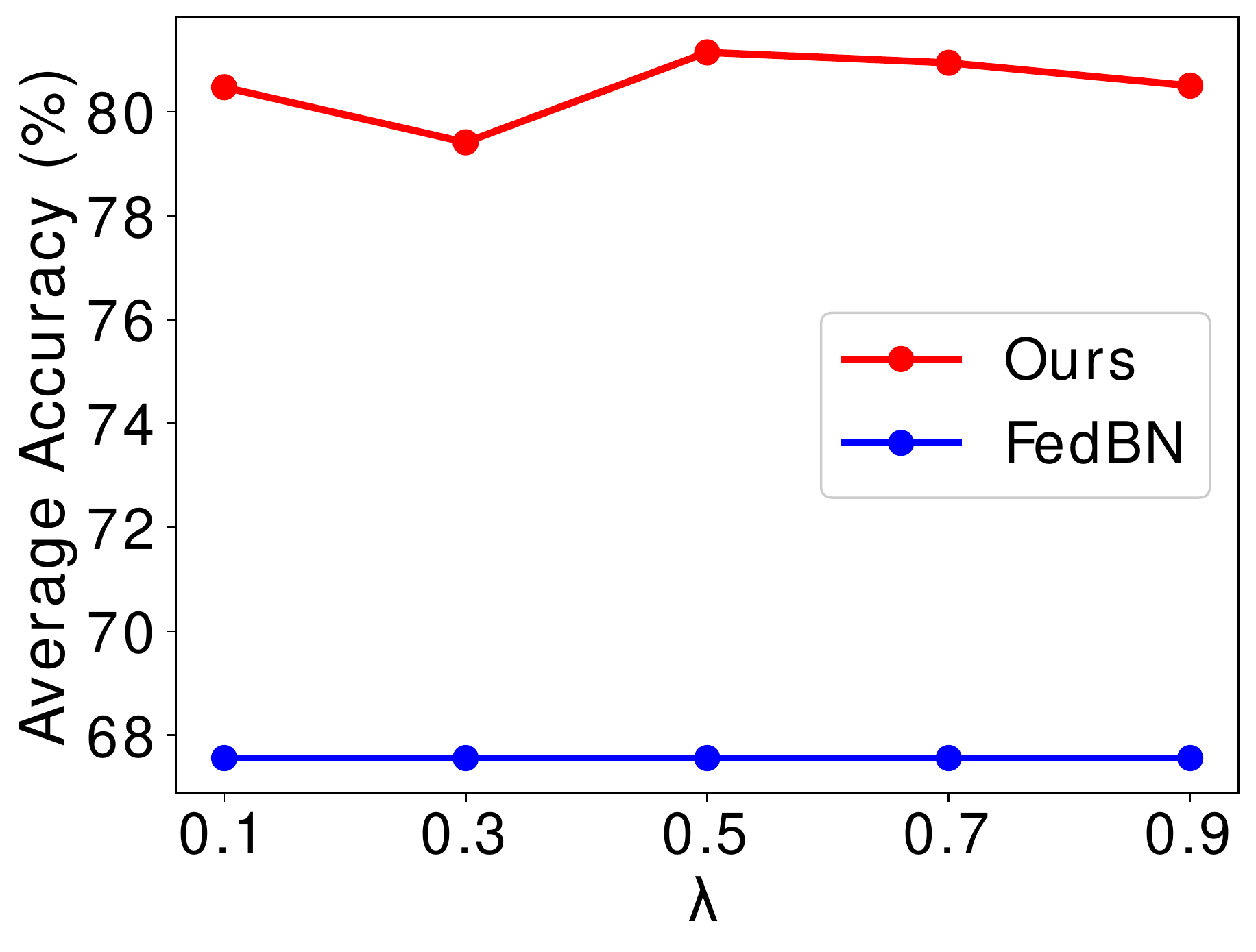}
	}
	\subfigure[$\lambda$]{
		\label{fig:s-lambdafun}
		\includegraphics[width=0.23\textwidth]{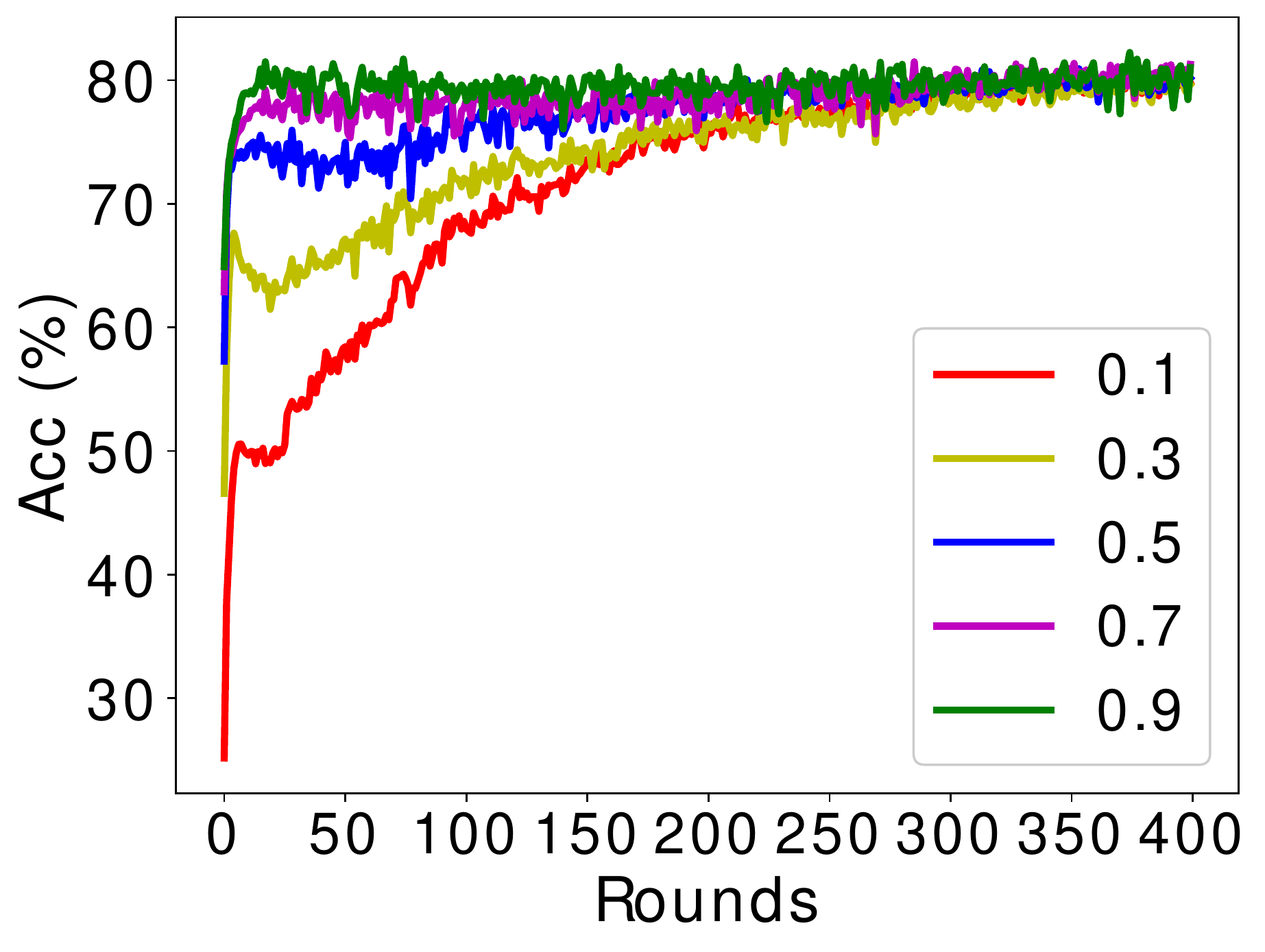}
	}
	\caption{Parameter sensitivity analysis.}
	\label{fig:sensity}
\end{figure*} 

\subsection{Ablation Study}

\textbf{Effects of Weighting.}

To demonstrate the effect of weighting which considers the similarities among the different clients, we compare the average accuracy on PAMAP2 and COVID-19 between the experiments with it and without it. Without weighting, our method degenerates to FedBN. From \figurename~\ref{fig:weffa}, we can see that our method performs much better than FedBN which does not include the weighting part. Moreover, from \figurename~\ref{fig:weff}, we can see our method performs better than FedBN on all clients. These results demonstrate that our method with weighting can cope with the label shifts while FedBN cannot deal with this situation, which means our method  is more applicable and effective. 

\noindent \textbf{Effects of Preserving Local Batch Normalization.}

We illustrate the importance of preserving local batch normalization. \figurename~\ref{fig:bneffa} shows the average accuracy between the experiments with preserving local batch normalization and the experiments with sharing common batch normalization while \figurename~\ref{fig:bneff} shows the results on each client. LBN means preserving local batch normalization while SBN means sharing common batch normalization. Obviously, the improvements are not particularly significant compared with weighting. This may be caused by there mainly exist the label shifts in our experiments while preserving local batch normalization is for the feature shifts. However, our method still has a slight improvement, indicating its superiority.\looseness=-1

\indent \textbf{Different Implementations of Our methods.}

In Method section, we propose three implementations of our method: \method, d-\method, and f-\method. The main differences among them are how to calculate $\mathbf{W}$. In \figurename~\ref{fig:p-other} and \figurename~\ref{fig:c-other}, we can see that all three implementations achieve better average accuracy on both PAMAP2 and COVID compared with FedAvg and FedBN. In addition, f-\method performs slightly worse than the other two variants, which may be because it only utilizes weighting during half rounds for fairness and the other half are for obtaining $\mathbf{W}$.

\begin{figure}[htbp]
    \centering
    \includegraphics[width=0.35\textwidth]{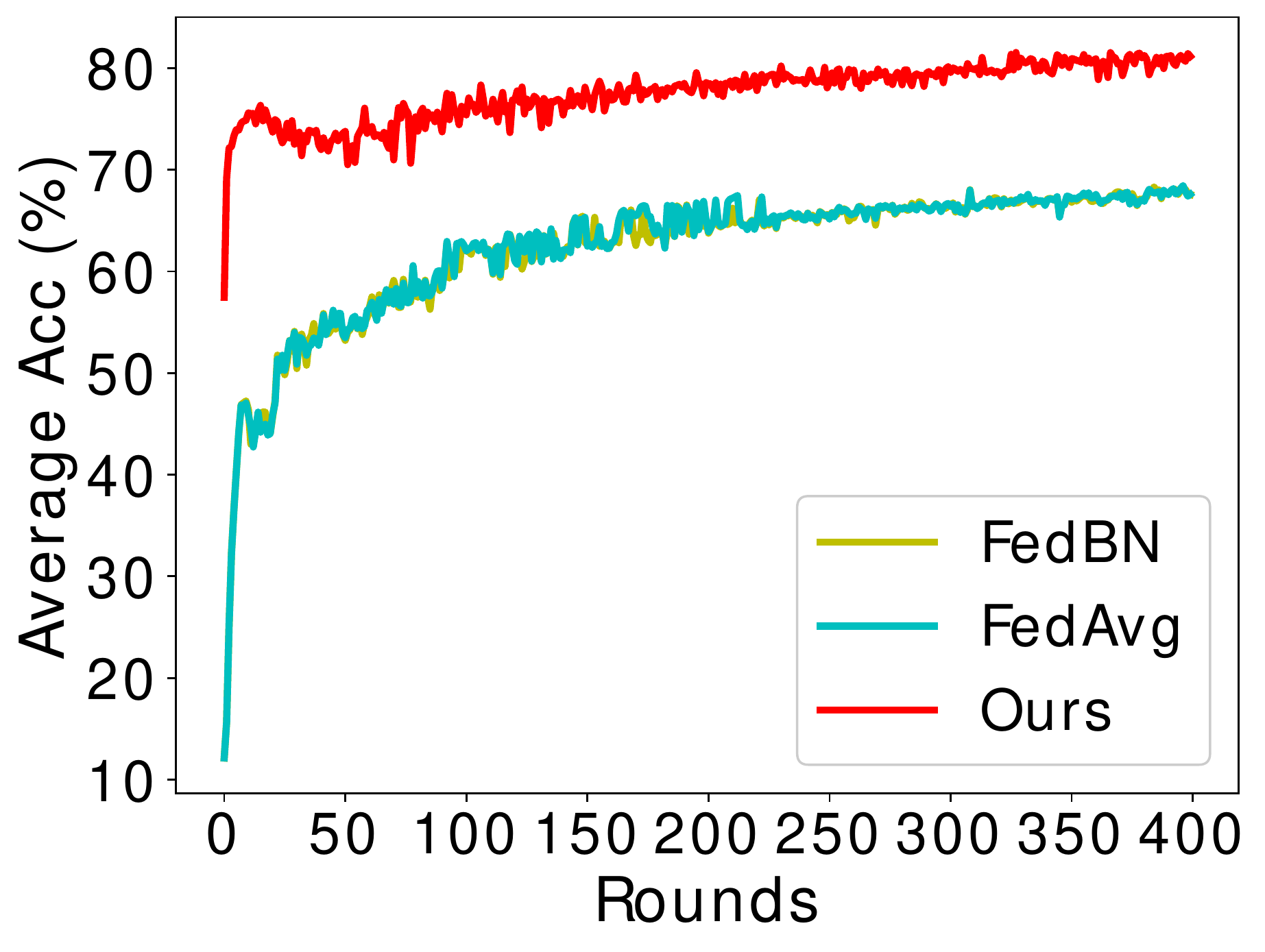}
    \caption{Convergence analysis of different methods.}
    \label{fig:convergence}
\end{figure}

\subsection{Convergence and Parameter Sensitivity}

We study the convergence of our method. From \figurename~\ref{fig:convergence}, we can see our method almost convergences in the $10$th round. And in the actual experiments, 20 rounds are enough for our method while FedBN needs over 400 rounds.

Then, we evaluate the parameter sensitivity of \method. Our method is affected by three parameters: local epochs, client number, and $\lambda$. We change one parameter and fix the other parameters. 

From \figurename~\ref{fig:s-clients}, we can see that our method still achieves acceptable results. 
When the client numbers increase, our method goes down which may be due to that few data in local clients make the weight estimation inaccurate. And we may take f-\method instead. 
In \figurename~\ref{fig:s-lepc}, we can see our method is the best and it is descending with local epochs increasing, which may be caused that we keep the total number of the epochs unchanged and the communication among the clients are insufficient. \figurename~\ref{fig:s-lambda}-\ref{fig:s-lambdafun} demonstrates $\lambda$ slightly affects the average accuracy of our method while it can change the convergence rate. The results reveal that \method is more effective and robust than other methods under different parameters in most cases.

\section{Conclusions and Future Work}

In this article, we proposed \method, a weighted personalized federated transfer learning algorithm via batch normalization for healthcare. \method aggregates the data
from different organizations without compromising privacy and security and achieves relatively personalized model learning through combing considering similarities and preserving local batch normalization. Experiments have evaluated the effectiveness of \method.
In the future, we plan to apply \method to more personalized and flexible healthcare. And we will consider better ways to calculate and update similarities among clients.

\section*{Acknowledgement}
This work is supported by the National Key Research and Development Plan of China No.2021YFC2501202, Natural Science Foundation of China (No. 61972383, No. 61902377, No. 61902379), Beijing Municipal Science \& Technology Commission No.Z211100002121171.

\ifCLASSOPTIONcaptionsoff
  \newpage
\fi

\bibliographystyle{IEEEtran}
\bibliography{IEEEabrv,mybibfile}

\end{document}